\documentclass[11pt]{article}

 \usepackage[final]{acl}
\usepackage{comment}
\usepackage{times}
\usepackage{latexsym}
\usepackage[utf8]{inputenc}
\usepackage{microtype}
\usepackage{amsmath,amssymb}
\usepackage{bm,bbm}
\usepackage{booktabs}
\usepackage{graphicx}
\usepackage{subcaption}
\usepackage{multirow}
\usepackage{enumitem}
\usepackage{xcolor}
\usepackage{listings}
\usepackage{algorithm}
\usepackage{algorithmic}
\usepackage{soul}

\title{
Biomedical Machine Translation for Low-Resource Arabic-Script Languages via Cross-Lingual Transfer and LoRA Adapter Merging
}

\author{
 \textbf{Abdullah Alabdullah \textsuperscript{1}},
 \textbf{Arash Eslamighayour \textsuperscript{1}},
 \textbf{Sarp Harbalioglu \textsuperscript{1}},
 \textbf{Lifeng Han\textsuperscript{2}},
\\
\\
 \textsuperscript{1}School of Informatics, University of Edinburgh, The United Kingdom\\
 \textsuperscript{2}Leids Universitair Medisch Centrum \& LIACS, Universiteit Leiden, NL\\
\\
 \small{
   \textbf{Correspondence:} \href{mailto:a.alabdullah@sms.ed.ac.uk}{a.alabdullah@sms.ed.ac.uk} \& \href{mailto:l.han@lumc.nl}{l.han@lumc.nl,liacs.leidenuniv.nl }
 }
}

\begin{document}
\maketitle

\begin{abstract} 
% \lipsum[1]  % Please comment out this line and start your own content.

% Here is how i usually write my abstracts:
% one short sentence for motivation and setting the scene
% what we did overall (overall achievement)
% What did we do 1 (first achievement) + briefly How did we do it + What did we find
% What did we do 2 (first achievement) + briefly How did we do it + What did we find
% What do we think it means (significance)

% Neural machine translation (NMT) for biomedical text remains largely inaccessible to hundreds of millions of speakers of underrepresented Arabic-script languages. We investigate whether healthcare-domain adaptation on two higher-resource pivot languages — Arabic and Persian — can support NMT for four severely low-resource languages: Dari, Pashto, Sorani Kurdish, and Urdu. Using Low-Rank Adaptation (LoRA) fine-tuning, we train strong pivot-language adapters and evaluate three cross-lingual transfer strategies: few-shot in-context learning, minimal supervised adaptation, and zero-data adapter merging. Supervised adaptation with just 500 sentences achieves near-pivot quality for Dari and meaningful gains for Urdu, while adapter merging provides a competitive zero-data alternative for Iranian-family languages. Pashto and Sorani Kurdish remain below deployable quality under all conditions, revealing the limits of cross-lingual transfer when structural distance from the pivot languages is too great.
%what about
We present a systematic study of healthcare-domain cross-lingual transfer to address the scarcity of biomedical NMT resources for Arabic-script languages.
%severe under-resourcing of biomedical NMT for Arabic-script languages. %using 
%what we did
We use Arabic and Persian as higher-resource pivots to improve translation for \textbf{four severely low-resource} targets: Dari (Afghan Persian, a standardised variety of Persian), Pashto, Sorani Kurdish (Central Kurdish, a major standardized variety of Kurdish), and Urdu (closely related to Hindi). Using LoRA fine-tuning on small decoder-only LLMs, we train \textit{domain-specific pivot adapters} and evaluate \textbf{three transfer strategies}: few-shot in-context learning, minimal supervised adaptation, and, to the best of our knowledge, for the first time in this setting, zero-data LoRA adapter merging. 
%result
Supervised adaptation with just 500 sentences achieves near pivot-language quality for Dari (CHrF++ 41.01) and meaningful gains for Urdu (28.88), while adapter merging reaches within 3.5 CHrF++ of supervised adaptation for Dari at zero additional cost. 
Pashto and Sorani Kurdish remain
insufficient for high-stakes clinical deployment
% below deployable quality across all strategies ($\leq 13$CHrF++), 
exposing the limits of cross-lingual transfer when structural distance from the pivots is too great. 
%impact/what we think
% Our results establish LoRA adapter merging as a cost-effective strategy for closely related Iranian-family languages.
LoRA adapter merging works surprisingly well for closely related languages, even without target-language biomedical data.

% LoRA adapter merging can provide a viable zero-resource biomedical adaptation strategy for closely related low-resource languages.
%Our results establish adapter merging as the most cost-effective strategy for Iranian-family languages.

\end{abstract} 

\section{Introduction}
% IF you like you can foollow this, if you think it is good/relevant
% - Motivation and context.
% - Gap in Existing Research/Literature Review (very brief just to state the need for our work).
% - Specific Problem Statement (the two problems or gaps I aim to address) + % - Brief overview of Contributions/Proposed Solution

\label{sec:intro}
Neural machine translation (NMT) has achieved remarkable progress for high-resource language pairs in general-domain settings \cite{johnson2017googles}. However, translating biomedical and clinical text remains a fundamentally harder challenge \cite{bawden-etal-2019-findings}. Unlike general-domain translation, where occasional inaccuracies cause minor miscommunication, errors in medical translation carry direct, life-threatening consequences \cite{mehandru2023physician}. Biomedical text is characterised by highly specialised terminology, context-sensitive acronyms (e.g., ``BP'' may refer to blood pressure or bipolar disorder), complex multi-word expressions, and a register that demands strict precision over fluency \cite{han2024neural}. These characteristics make biomedical NMT qualitatively distinct from general-domain MT and require dedicated system development, domain adaptation, and safety-aware evaluation.

Despite growing reliance on AI-based MT in healthcare for translating discharge instructions, facilitating doctor–patient communication, and supporting clinical decision-making \cite{zappatore2024adopting}, the vast majority of biomedical MT research has concentrated on well-represented European languages \cite{neves2024findings, han2024neural}. 
Arabic-script languages spoken across the Middle East and Central Asia remain critically under-studied in this domain \cite{abouzahir-etal-2026-cross}, despite collectively serving hundreds of millions of speakers in contexts where interpreters are often unavailable and mistranslation carries direct clinical consequences \cite{al_shamsi_implications_2020}.

A central obstacle is the lack of domain-specific parallel data for these languages \cite{nigatu-etal-2025-viability,merx2025openwho}. While multilingual models such as NLLB-200 provide broad language coverage, they are trained on general-domain data and require domain-specific adaptation to perform reliably in healthcare settings \cite{nllbteam2022languageleftbehindscaling_arXiv,han2023investigating}. The large parallel corpora needed for such adaptation are typically unavailable for low-resource languages, motivating transfer-based approaches that leverage knowledge from related higher-resource languages --- an approach supported by evidence of positive cross-lingual transfer under joint training \cite{lakew2018comparison, aharoni2019massively}.

In this work, we investigate transfer from two higher-resource pivot languages~\cite{talwar2025pivotlanguagelowresourcemachine} (Arabic (ar) and Persian (fa)) to four severely low-resource targets: 1) Dari (prs), a standard variety of Persian; 2) Pashto (ps), an Eastern Iranian language ; 3) Sorani Kurdish (ckb), a major dialect of Kurdish; and 4) Urdu (ur), closely related to Hindi language. All four languages have their scripts written right-to-left with some comparisons listed in Table \ref{tab:nlp_language_comparison_compact}.\footnote{Pashto: written in a modified Perso-Arabic script, as one of the official languages of Afghanistan and widely spoken in Pakistan, overall 40-60 million speakers \url{https://www.britannica.com/topic/Pashto-language}.
}
Dari shares substantial lexical, grammatical, and
orthographic overlap with Persian. Pashto and Sorani
show greater morphological and orthographic divergence,
while Urdu benefits primarily from script overlap and
extensive Persian/Arabic lexical borrowing.
Both pivots have sufficient publicly-available biomedical parallel data for meaningful fine-tuning, and together provide complementary transfer signals: Arabic shares script, vocabulary, and loanwords with all four target languages, while Persian is closely related to Dari (a mutually intelligible variety), shares script and vocabulary with Pashto and Sorani Kurdish, and has substantially influenced Urdu at the lexical level. The four target languages are selected for their extreme data scarcity, limited MT research coverage, and linguistic proximity to the pivots.

% This motivates our central research question (\textbf{RQ}): {\textit{to what extent can domain adaptation on the higher-resource pivot languages Arabic and Persian support healthcare-domain MT for related low-resource languages through cross-lingual transfer?}}
Our \textbf{research questions} are: RQ1)
How effective is biomedical domain transfer from Arabic and Persian to related low-resource Arabic-script languages?
RQ2)
Can zero-shot LoRA adapter merging approximate supervised adaptation?
RQ3)
How does linguistic distance influence transfer effectiveness?
To investigate these, we first fine-tune LoRA adapters \cite{hu2022lora} on healthcare-domain data for English-to-Arabic (\texttt{en}$\rightarrow$\texttt{ar}) and English-to-Persian (\texttt{en}$\rightarrow$\texttt{fa}) translation, then examine whether the domain knowledge encoded in these pivot adapters transfers to four low-resource target languages through three complementary cross-lingual transfer strategies. This paper makes the following contributions and interesting findings:
\begin{itemize}[noitemsep, topsep=0pt]
    \item We present the first systematic evaluation of LoRA adapter merging via tensor arithmetic for biomedical NMT across Arabic-script low-resource languages.
    \item We show that supervised adaptation with just 500 sentences achieves pivot-level quality for Dari (CHrF++ 41.01), demonstrating the viability of cross-lingual transfer under severe data constraints.
    \item We demonstrate that LoRA adapter merging enables zero-resource biomedical domain adaptation for closely related low-resource Arabic-script languages, achieving performance surprisingly close to supervised fine-tuning while requiring no target-language biomedical parallel data.
    \item We have an interesting finding of a model inversion effect: a weaker pivot model (Llama~3.2~3B) transfers more effectively to low-resource targets than a stronger one (Gemma~2~2B), suggesting a possible trade-off between pivot specialisation and cross-lingual generalisation.
    % \item We identify a model inversion effect: a weaker pivot model (Llama~3.2~3B) transfers more effectively to low-resource targets than a stronger one (Gemma~2~2B), revealing a trade-off between pivot specialisation and cross-lingual generalisation.

\end{itemize}
The rest of the paper is organised as below: Section~\ref{sec:background} reviews related work; Section~\ref{sec:dataset_task} describes the data and task; Section~\ref{sec:method} presents the methodology; Section~\ref{sec:exps} reports experiments and results, and Section~\ref{sec:conc} concludes.

% \begin{table}[th!]
% \centering
% \tiny 
% \setlength{\tabcolsep}{3.5pt}
% \begin{tabular}{p{1.4cm} p{1.9cm} p{1.3cm} p{1.7cm} p{2.3cm} p{1.6cm}}
% \toprule
% \textbf{Lang.} & \textbf{Family} & \textbf{Script} & \textbf{Persian Similarity} & \textbf{Key NLP Factors} & \textbf{Transfer Difficulty} \\
% \midrule
% Dari & Western Iranian (Persian) & Perso-Arabic & Very high & Strong lexical, grammatical, and orthographic overlap & Low \\
% Pashto & Eastern Iranian & Modified Perso-Arabic & Low--moderate & Shared script but substantial lexical/morphological divergence & High \\
% Sorani & Northwestern Iranian & Sorani Arabic-based & Moderate & Partial overlap, but differing morphology and orthography & Moderate--high \\
% Urdu & Indo-Aryan & Persian-based Arabic & Low genealogically & Script overlap and heavy Persian/Arabic borrowing, but distinct grammar & Moderate \\
% \bottomrule
% \end{tabular}
% \caption{NLP-oriented summary of the target languages.}
% \label{tab:nlp_language_comparison_compact}
% \end{table}

\begin{table}[t]
\centering
\tiny 
\begin{tabular}{lcccc}
\toprule
\textbf{Language} &
\textbf{Family} &
\textbf{Script} &
\textbf{Persian Sim.} &
\textbf{Difficulty} \\
\midrule
Dari   & W. Iranian  & Shared  & Very High & Low \\
Pashto & E. Iranian  & Shared  & Low--Mod. & High \\
Sorani & NW. Iranian & Partial & Moderate  & Mod.--High \\
Urdu   & Indo-Aryan  & Shared  & Low       & Moderate \\
\bottomrule
\end{tabular}
\caption{
Expected transfer difficulty based on linguistic
relationship to the pivot languages.
}
\label{tab:nlp_language_comparison_compact}
\end{table}

\section{Related Work}
\label{sec:background}
% Machine translation (MT), the task of automatically converting text from one 
% natural language to another, represents one of the longest-standing and most 
% challenging problems in artificial intelligence. Its history spans more than 
% seven decades, evolving from rigid rule-based systems to statistical frameworks and, ultimately, to the data-driven neural paradigm that dominates modern research
% seven decades, evolving from rigid rule-based systems to statistical frameworks and, ultimately, to the data-driven neural paradigm that dominates modern research

This work sits at the intersection of three bodies of work: LLMs for low-resource translation, cross-lingual transfer from high-resource pivot languages, and parameter-efficient adapter composition. We review each in turn.

% \todo{consider adding more biomed NMT?}
\subsection{LLMs for low-resource and biomedical translation}
NMT has traditionally relied on encoder--decoder architectures \citep{bahdanau2015neural, sutskever2014sequence}, but decoder-only LLMs have recently demonstrated competitive performance with less parallel data \citep{xu2024paradigm}. However, their effectiveness remains limited for low-resource languages: \citet{zhu2024multilingual} evaluated LLMs across 102 languages and found consistently poor performance on low-resource directions. \citet{mao-yu-2024-tuning} address this with contrastive alignment instructions for unseen languages, while \citet{xiao2025lsftl} show that language-specific LoRA fine-tuning substantially improves LLM-based MT for low-resource settings. In the medical domain, conventional MT systems still outperform LLMs on complex clinical texts \citep{li2026comparing}, motivating our strategy of fine-tuning domain-specific LoRA adapters rather than relying on zero-shot LLM capabilities.

\subsection{Cross-lingual transfer}
% Cross-lingual transfer leverages knowledge learned from high-resource languages to improve performance on low-resource languages for which labelled data is scarce. The effectiveness of such transfer depends on morphological similarity and vocabulary overlap \citep{conneau2020unsupervised, aharoni2019massively}. \citet{garcia2023unreasonable} showed that few-shot in-context learning can match fine-tuned baselines when the base model is sufficiently capable. For Arabic-script languages, shared script aids subword-level transfer, though character-level variation across languages weakens this advantage \citep{muller2021interplay}. 

Cross-lingual transfer leverages knowledge from high-resource languages to improve low-resource performance, with effectiveness depending on linguistic similarity \citep{conneau2020unsupervised, aharoni2019massively}. \citet{chronopoulou2023language} show that language-family adapters outperform language-agnostic ones, suggesting typologically proximate pivot languages yield stronger transfer. For Arabic-script languages, shared script aids subword-level transfer, though character-level variation weakens this advantage \citep{muller2021interplay}. \citet{garcia2023unreasonable} show that few-shot in-context learning can match fine-tuned baselines when the base model is sufficiently capable. In our setting, Arabic and Persian serve as pivot languages for transfer to Dari, Pashto, Sorani Kurdish, and Urdu.

\subsection{Adapter merging for cross-lingual transfer}
Parameters of independently fine-tuned models can be combined through tensor arithmetic on model weights \citep{wortsman2022model_soups, ilharco2023editing}. These approaches have been used primarily in multi-task settings, where models trained on different tasks are merged to produce a single model with combined capabilities.

% More recently, work has extended this idea to adapter-level merging and cross-lingual transfer. \citet{zhao2025adamergex} decompose cross-lingual ability into task and language adapters and merge them for NLU on LLaMA-2. \citet{bandarkar2025unreasonable} show that merging language and task experts via layer-swapping is effective for cross-lingual reasoning. \citet{li2024unlocking} demonstrate that merging continually pre-trained models with English task models outperforms standard fine-tuning when target-language data is scarce. However, none of these works merge domain-specific LoRA adapters trained on different languages to construct a single multilingual adapter for translation. This is the gap we address in this work.

More recently, work has extended this idea to adapter-level merging and cross-lingual transfer. \citet{zhao2025adamergex} decompose cross-lingual ability into task and language adapters and merge them for NLU. \citet{bandarkar2025unreasonable} show that merging language and task experts via layer-swapping is effective for cross-lingual reasoning. \citet{tao2024unlocking} demonstrate that merging continually pre-trained models with English task models outperforms standard fine-tuning when target-language data is scarce. However, none of these works merge \emph{domain-specific} LoRA adapters trained on \emph{different languages} to construct a single multilingual adapter for translation in specialised domains. This is the gap we address in this work.

\section{Methodology}
\label{sec:method}

\subsection{Fine-tuning with Low-Rank Adaptation}

All models used in our study are autoregressive decoder-only transformers, which generate each token conditioned on all previously generated tokens and the source input. All six models share this architecture but differ in scale (1.5B--3B parameters), vocabulary size, and multilingual pretraining data.

We apply LoRA to fine-tune domain-specific adapters for English-to-Arabic and English-to-Persian translation, forming the foundation for all subsequent transfer experiments. Large pre-trained language models contain billions of parameters, making full fine-tuning computationally expensive. Low-Rank Adaptation (LoRA)~\cite{hu2022lora} addresses this by freezing the original weights and injecting small trainable matrices into selected layers. For a weight matrix $\mathbf{W}_0 \in \mathbb{R}^{d \times k}$, LoRA defines the update as:
\begin{equation}
    \mathbf{W} = \mathbf{W}_0 + \mathbf{B}\mathbf{A},
    \label{eq:lora}
\end{equation}
where $\mathbf{B} \in \mathbb{R}^{d \times r}$ and $\mathbf{A} \in \mathbb{R}^{r \times k}$ are the only trainable parameters, with rank $r \ll \min(d, k)$. The update is scaled by $\alpha/r$, where $\alpha$ controls its magnitude. Only $\mathbf{A}$ and $\mathbf{B}$ are trained, allowing trainable parameter reduction by up to 99\% while preserving pre-trained knowledge. The resulting adapter can be stored and applied independently of the base model, with each adapted layer adding approximately $2rd$ parameters.

\subsection{Multilingual Adapter Training}
\label{subsec:joint-training}

We use joint multilingual training to produce a single Arabic--Persian adapter, serving as an upper bound reference for our adapter merging approach. Rather than training separate adapters per language, joint training learns a single adapter from a shuffled multilingual dataset, with a language-conditioned prompt specifying the target language per input. Since all languages share the same adapter parameters, they compete for limited representational capacity; a rank that is too small risks interference, where gains in one language degrade the other.

% Meanwhile, increasing the rank expands the representational capacity, but at the cost of higher memory usage during training and larger adapter checkpoints at inference.

\subsection{Zero-Shot and Few-Shot Transfer Learning}

We apply the pivot-language adapters directly at inference time to evaluate how much domain knowledge transfers to low-resource target languages without any additional training.

Transfer learning leverages knowledge from a source language to improve performance on a target language with limited labelled data~\cite{pan2010survey}. In our setting, we explore two inference-time strategies that require no additional training on target-language data.

In zero-shot transfer, the Arabic and Persian adapters are applied directly to Dari, Sorani Kurdish, Pashto, and Urdu without any target-language examples. This is most effective when languages share script, morphology, or vocabulary~\cite{aharoni2019massively, conneau2020unsupervised}, all of which hold to varying degrees across our target languages.

In few-shot in-context learning (ICL)~\cite{brown2020gpt3}, $k$ target-language translation examples are prepended to the input at inference time. This steers the model toward the target language without updating any parameters, and has been shown to rival fine-tuned models when the base model is sufficiently capable~\cite{garcia2023unreasonable}.

\subsection{Model Merging for Parameter-Efficient Adapters}
\label{subsec:model-merging}

We merge the Arabic and Persian LoRA adapters via tensor arithmetic to produce a zero-data multilingual adapter for transfer to low-resource target languages.

Model merging combines the capabilities of multiple independently fine-tuned models into a single model without additional training \cite{wortsman2022model_soups, ilharco2023editing}. Rather than retraining on a joint dataset, it operates directly in weight space by combining model parameters through tensor arithmetic. When the models are sufficiently aligned, the resulting model can retain the competencies of each constituent \cite{ilharco2023editing}.

This approach has negligible computational cost and works best when models share the same base architecture and initialisation, ensuring aligned parameter spaces for meaningful element-wise operations \citep{yadav2023resolving}. Merging can be applied to full models or to adapters (e.g. LoRA matrices). In the latter case, adapters trained from a shared frozen backbone lie in a common parameter space, enabling their weights to be combined directly \citep{hu2022lora}.

Formally, let each LoRA adapter be indexed by $i \in \{1, 2\}$. Let $\boldsymbol{\theta}_0$ denote the full parameter vector of the frozen base model, and $\boldsymbol{\theta}_i$ denote the effective parameters after applying adapter $i$. Since the base weights are frozen during LoRA training, the only learned parameters are the adapter matrices, and the difference $\boldsymbol{\tau}_i = \boldsymbol{\theta}_i - \boldsymbol{\theta}_0$, referred to as the \emph{task vector}~\citep{ilharco2023editing}, is entirely determined by the adapter. Because both adapters are trained from the same frozen base with identical LoRA configurations, they share the same tensor keys and shapes, making element-wise operations on their parameters directly applicable. The merged adapter is then obtained by combining the two task vectors and adding them back to the base:
\begin{equation}
    \boldsymbol{\theta}_{\text{merged}} = \boldsymbol{\theta}_0 + f(\boldsymbol{\tau}_1, \boldsymbol{\tau}_2),
\end{equation}
where $f$ denotes a merging function applied to the task vectors. The specific choices of $f$ are described in the following subsections.
We describe more details of weighted averaging, TIES-Merging, and DARE (Drop and rescale) in Section \ref{appendix:model-merge-details}.

\section{Dataset and Task Formalisation}
\label{sec:dataset_task}

\subsection{Task Formulation}
We address sentence-level, domain-specific NMT from English (en) into six target languages: Arabic (ar), Persian (fa), Dari (prs), Sorani Kurdish (ckb), Pashto (ps), and Urdu (ur). Arabic and Persian serve as \emph{pivot languages} with available in-domain training data; Dari, Sorani Kurdish, Pashto, and Urdu are \emph{low-resource target languages} for which no large-scale healthcare-domain parallel data exists. This data constraint motivates a transfer learning framework in which domain knowledge encoded by pivot-language adapters is redirected to typologically related target languages. Concretely, our task decomposes into three sub-problems: (i) fine-tuning domain-specific LoRA adapters for ar and fa, (ii) transferring these adapters to the four low-resource target languages via zero-shot or few-shot inference, and (iii) investigating whether minimal target-language supervision or zero-data adapter merging can further improve transfer quality.

\subsection{Data}
\subsubsection{Training Data}
\label{trainingdata}
As no single healthcare-domain parallel dataset covers both en$\rightarrow$ar and en$\rightarrow$fa, we use two separate corpora. For en$\rightarrow$ar, we use PEACH~\cite{peach2023}, a professionally translated, sentence-aligned corpus of patient information leaflets and patient education materials. PEACH's patient-facing content (e.g., instructional sentences) aligns closely with our goal of translating public health communication. The dataset is manually aligned, ensuring high quality.

For en$\rightarrow$fa, we use the Medical Science subset of Esposito~\cite{esalati2024esposito}, consisting of sentence-aligned pairs from bilingual scientific journal abstracts. While its content is more scientific, its medical vocabulary remains relevant to healthcare text translation. Unlike PEACH, Esposito is automatically aligned; however, human evaluation of 1{,}500 samples shows that approximately 82\% of sentence pairs are semantically well aligned (``Good'' or ``Very Good'' labelled). It has also been validated for MT, with models trained on it outperforming general-domain baselines~\cite{esalati2024esposito}.

Both datasets were preprocessed using sentence-length filtering, retaining pairs with source lengths between 5 and 80 words. This removes short fragments with limited translation value (e.g., headers) and long outliers likely caused by alignment errors — a standard filtering criterion in NMT preprocessing~\cite{koehn-etal-2007-moses}. 
After filtering, only 25{,}000 sentence pairs were sampled from each corpus for computational feasibility and shuffled to remove document-ordering effects. All 25{,}000 pairs from each corpus are used for training; no held-out split is taken from the training corpora.

For our four low-resource target languages, no healthcare-domain parallel data exists. We instead sample 500 sentence pairs per language from FLORES-200~\cite{nllbteam2022languageleftbehindscaling_arXiv} — a professionally translated, general-domain multilingual corpus derived from Wikipedia and Wikinews — covering Dari (prs), Sorani Kurdish (ckb), Pashto (ps), and Urdu (ur). These small general-domain sets are used exclusively for two purposes: as source demonstrations in few-shot in-context learning (\S\ref{sec:fewshot}), and as fine-tuning data in our adaptation experiment (\S\ref{sec:adapt_ft}).

\subsubsection{Evaluation and Test Data}
For evaluation, we use the TICO-19 (Translation Initiative for COvid-19) benchmark~\cite{tico}, a professionally translated multilingual dataset comprising healthcare and public health sentences covering all six languages in our study. We use the provided split: 971 sentences for intermediate evaluation and model selection, and 2,100 for final testing. The dataset spans patient-facing guidance, scientific text, and general public health content, making it both domain-relevant and a realistic benchmark relative to our training data. No preprocessing was required for either sets.

TICO-19's high-quality translation process, widespread use in low-resource MT research, and cross-lingual comparability (due to shared English source sentences) make it well-suited to this study. We use TICO-19 exclusively for evaluation and testing, ensuring no data contamination with training corpora.

\subsection{Evaluation Setup}

Following the recent Arabic NMT work from \newcite{alabdullah2025advancing},
we use CHrF++ \cite{popovic-2017-chrf} as our primary evaluation metric. CHrF++ computes an F-score over character n-grams (up to length 6) augmented with word unigram and bigram overlap, weighted toward recall by default. This design makes it particularly well-suited to our setting for three reasons. First, character-level overlap awards partial credit to morphological variants that differ only in prefix, suffix, or inflection --- a critical advantage for Arabic and Persian, where a single lemma can surface in dozens of inflected forms that exact word matching would treat as entirely incorrect \cite{chauhan2021adableu}. Second, its recall orientation penalises translations that omit content, not only those that produce incorrect words --- important in a medical domain where missing information can be as dangerous as mistranslation. Third, CHrF++ has been adopted as the primary metric in recent low-resource MT shared tasks precisely because of its robustness to morphological richness \cite{popovic-2015-chrf}, making our results directly comparable to the broader low-resource MT literature. BLEU \cite{10.3115/1073083.1073135} is reported in Appendix B but not used as primary metric: its exact word-level matching systematically penalises valid morphological variants, a well-documented limitation for inflectionally rich languages \cite{chauhan2021adableu,popovic-2017-chrf}. COMET \cite{rei-etal-2020-comet} is similarly relegated to the Appendix, as it is trained predominantly on European-language judgements and offers reduced reliability for our Arabic-script targets \cite{rei-etal-2020-comet}. All CHrF++ scores are computed using \texttt{sacrebleu} on the 0--100 scale.

\textbf{{Text Normalisation}}
Arabic-script languages often encode the same character via multiple Unicode representations \cite{inproceedings}, which can cause metrics to penalise valid translations. We therefore apply normalisation to both hypothesis and reference at evaluation time: CAMeL Tools~\cite{obeid-etal-2020-camel} for Arabic, and Hazm~\cite{hazm} for Persian and Dari. No normalisation is applied to Pashto, Sorani Kurdish, or Urdu, as no standardised tools exist for these languages.

We investigate our research question (\S\ref{sec:intro}) using a six-stage experimental pipeline that progressively builds toward our cross-lingual transfer goal. We first describe the configurations used across experiments (\S\ref{sec:exp_setup}), followed by model selection, and monolingual and joint multilingual fine-tuning (\S\ref{sec:pivot_training}). We then introduce three complementary transfer strategies: few-shot in-context learning (\S\ref{sec:fewshot}), adaptation fine-tuning (\S\ref{sec:adapt_ft}), and zero-data adapter merging (\S\ref{sec:adapter_merge}). 
For each experiment, we explain its motivation and relevance to the research objective, followed by a technical description of the experimental procedure.
We describe such details on experimental setup, pilot language adapter training, few-shot cross-lingual transfer, adaptation fine-tuning on low-resource target languages, and LoRA adapter merging for zero-data cross-lingual transfer in Section \ref{sec:exps} for reproducibility.

\section{Results and Discussion}
\label{sec:results}

\begin{table}[h]
\centering
\resizebox{\columnwidth}{!}{%
\begin{tabular}{lrrrr}
\toprule
\textbf{Configuration} & \textbf{PRS} & \textbf{CKB} & \textbf{PS} & \textbf{UR} \\
\midrule
Zero-shot (Gemma)             &  5.44 &  3.23 &  2.77 & 13.40 \\
Zero-shot (Llama)             & 12.64 &  3.10 &  2.56 &  4.08 \\
\midrule
Few-shot (AR)                & 25.94 &  8.93 &  6.10 & 12.25 \\
Few-shot (FA)                & 14.64 &  8.83 &  6.57 &  8.16 \\
Few-shot (AR+FA) (Llama)       & 34.84 & 15.42 & 16.70 & 22.26 \\
\midrule
Adaptation (AR)              & \textbf{41.01} & 12.34 &  9.94 & \textbf{28.88} \\
Adaptation (FA)              & 40.25 & \textbf{12.97} & 10.31 & 28.74 \\
\midrule
Merge: Simple Avg            & 37.56 & 10.03 & 12.23 & 15.67 \\
Merge: Wtd.\ ($\alpha{=}0.3$) & 36.50 & 11.34 & \textbf{12.87} & 13.94 \\
Merge: Wtd.\ ($\alpha{=}0.7$) & 36.12 &  7.04 & 10.53 & 16.69 \\
Merge: TIES                  & 36.49 &  7.67 &  9.03 & 15.05 \\
Merge: DARE                  & 37.43 &  9.91 & 11.99 & 15.26 \\
\bottomrule
\end{tabular}}
\caption{CHrF++ on four low-resource target languages. Rows group into zero-shot, few-shot transfer, adaptation ($\sim$500 sentences), and adapter merging. \textbf{Bold}: best per column. Unless otherwise specified the model used is Gemma}
\label{tab:main-results}
\end{table}

\subsection{Model Selection}
Gemma 2 2B and Llama 3.2 3B produce meaningful zero-shot outputs on Arabic and Persian and improve with fine-tuning, whereas BLOOM 1.7B, XGLM 1.7B, and mGPT 1.3B generate near-random outputs with negligible gains (e.g., BLOOM Arabic CHrF++ 5.86$\rightarrow$5.94). Despite their multilingual design, the latter models fail entirely, indicating that at this scale, the quantity and quality of Arabic-script pretraining data matter more than multilinguality alone.

Gemma 2 2B achieves the highest zero-shot scores on Arabic (29.91) and Persian (28.49), rising to 33.97 and 40.09 after fine-tuning on 500 examples, suggesting strong Arabic and Persian pretraining. Llama 3.2 3B shows similar behaviour, improving from 13.97$\rightarrow$26.97 (Arabic) and 10.66$\rightarrow$30.58 (Persian) with minimal data. All models achieve near-zero CHrF++ on low-resource languages (Dari, Sorani Kurdish, Pashto, \& Urdu), confirming the need for dedicated adaptation and motivating transfer-learning experiments. Accordingly, Gemma 2 2B and Llama 3.2 3B are selected for subsequent experiments.
\subsection{Pivot Language Baselines}
Scaling pivot-language training from 500 to 25,000 sentences yields strong monolingual adapters for Gemma 2 2B, with asymmetric gains: Arabic improves substantially (33.97$\rightarrow$40.42), while Persian remains unchanged (40.09$\rightarrow$40.06), saturating after 500 examples. This likely reflects strong Persian pretraining (requiring minimal domain adaptation) and limited lexical diversity in the Esposito corpus, reducing returns from additional data. Arabic, by contrast, continues to benefit from scaling, suggesting our Arabic fine-tuning data is more distant from Gemma's priors. Llama 3.2 3B scales more uniformly (AR: 26.97$\rightarrow$30.48; FA: 30.58$\rightarrow$35.15) but reaches lower ceilings on both languages.

Joint AR+FA fine-tuning shows that both languages can be encoded in a single adapter without interference: Gemma improves substantially above its monolingual baselines (AR: 40.42$\rightarrow$42.45; FA: 40.06$\rightarrow$42.20), establishing the strongest pivot baselines. Llama, by contrast, exhibits cross-lingual interference under joint training, with both languages degrading below monolingual performance (AR: 30.48$\rightarrow$28.87; FA: 35.15$\rightarrow$31.44), indicating that the capacity for positive joint transfer is model-dependent.

\subsection{Few-Shot Transfer Learning}
Pivot-language adapters allow transfer to all four unseen target languages, with quality closely tracking linguistic proximity. Three in-context examples raise Dari well above zero-shot (Gemma AR: 16.33$\rightarrow$25.94), demonstrating that inference-time adaptation yields large gains at zero cost. Performance varies sharply: Dari achieves the highest scores (Gemma AR adapter: 25.94; Llama AR+FA adapter: 34.84), while Sorani Kurdish, Pashto, and Urdu remain much lower (6--22), consistent with greater structural distance.

A key finding is a model inversion: although Gemma outperforms Llama on pivot languages (40--42 vs.\ 28--35), Llama transfers better than Gemma across all low-resource languages, especially Pashto (16.70 vs.\ 6.28) and Urdu (22.26 vs.\ 9.96). 
% This reflects a trade-off between pivot specialisation and cross-lingual generalisation:
One possible explanation is a trade-off between pivot specialisation and cross-lingual generalisation:
intensive LoRA fine-tuning on Arabic and Persian overwrites Gemma’s broader multilingual representations, reducing transfer capacity, whereas Llama retains more generalisable cross-lingual priors.

Adapter source reveals a further asymmetry: under Gemma, the Arabic adapter outperforms the Persian adapter for Dari (25.94 vs.\ 14.64) despite closer linguistic similarity. Under Llama, the expected pattern holds (FA: 32.87 $>$ AR: 31.33), indicating model-specific interactions between pivot-language adapter training and transfer. Across both models, the combined AR+FA adapter consistently matches or exceeds the best single adapter.

\subsection{Adaptation with Minimal Supervision}
Minimal target-language fine-tuning (with AR and FA adapters) is highly effective for Dari and Urdu but exposes a performance ceiling for Pashto and Sorani Kurdish that $\sim$500 sentences cannot overcome. Dari reaches 41.01 (AR adapter) and 40.25 (FA adapter), a +15.1 gain over the best few-shot result (25.94), matching Arabic and Persian pivot baselines (40.42 and 40.06). The few-shot advantage of the AR adapter (25.94 vs.\ 14.64) disappears after adaptation, with both adapters converging, consistent with Dari’s status as a Persian variety. This suggests that without target-language data, corpus quality drives transfer, whereas with supervision, linguistic proximity dominates. Urdu shows even larger gains (+16.6), reaching 28.88 (AR) and 28.74 (FA), indicating that shared script and lexical borrowing enable rapid adaptation despite its Indo-Aryan typology.

In contrast, Pashto and Sorani Kurdish improve only +3.4--4.2 CHrF++ and remain below 13, far from usable performance. Pashto (FA: 10.31; AR: 9.94) diverges in morphosyntax and script, while Sorani Kurdish (FA: 12.97; AR: 12.34) shares vocabulary but differs grammatically, gaps that 500 general-domain sentences cannot bridge. Addressing this likely requires more data, in-domain supervision, or a closer pivot language. Across all targets, AR and FA adapters converge within 1.0 CHrF++, indicating that pivot choice becomes irrelevant once target-language supervision is available.

\subsection{Adapter Merging as a Zero-Data Strategy}
Merging AR and FA adapters yields asymmetric effects across pivot languages. On the Persian dev set, weighted averaging ($\alpha=0.7$) achieves 42.54, the best Persian result in the study, surpassing both the AR+FA jointly-trained adapter (42.20) and the monolingual FA adapter (40.06) at zero additional cost. This suggests the AR adapter contributes complementary Arabic-script features (e.g., shared vocabulary and biomedical terminology) that enhance Persian performance. The reverse does not hold: on Arabic dev set, all merging methods degrade performance, with the best result ($\alpha=0.7$, 39.95) falling 0.47 below the monolingual AR adapter and 2.5 below AR+FA jointly-trained adapter. Practically, merging provides a cost-free improvement for Persian outperforming even joint training where strong single-language adapters exist.

Transfer to low-resource targets shows strong language-dependent variation. Dari reaches 37.56 CHrF++ (Simple Average), within 3.5 points of adaptation (41.01) despite using no Dari data, and far exceeding few-shot (25.94), making it a strong zero-data model. For Pashto, weighted averaging ($\alpha=0.3$) achieves 12.87, outperforming adaptation (10.31) by +2.56 — a rare case where zero-data merging surpasses supervised fine-tuning. This may reflect domain mismatch (where our domain-agnostic adaptation data is shifting adapters away from biomedical domain) or insufficient data to capture Pashto’s structural diversity, though confirming this requires further ablation. Urdu shows the opposite pattern: the best merging result (16.69, $\alpha=0.7$) lags far behind adaptation (28.88; $-12.2$), indicating that weight-space interpolation cannot replace in-language supervision under high linguistic distance. Sorani Kurdish lies between these extremes, with merging (11.34, $\alpha=0.3$) trailing adaptation (12.97) by only 1.6, though neither reaches usable performance.

Across target languages, optimal merging tracks linguistic proximity to the pivots: Persian-biased weighting ($\alpha=0.3$) performs best for Iranian-family targets (Dari, Pashto, Sorani Kurdish), while Arabic-biased weighting ($\alpha=0.7$) is preferred for Urdu and the pivot languages. TIES ($k=0.2$) consistently underperforms, as trimming 80\% of LoRA parameters removes shared signal rather than noise. Table 1 shows the full comparison; overall, merging is the most cost-effective starting point for Iranian-family targets without parallel data, whereas Urdu requires in-language supervision.

\subsection{Test Set Evaluation}
\label{sec:test_res}

\begin{table}[t]
\centering
\resizebox{\columnwidth}{!}{%
\begin{tabular}{lrrrrrr}
\toprule
\textbf{Configuration} & \textbf{AR} & \textbf{FA} & \textbf{PRS} & \textbf{CKB} & \textbf{PS} & \textbf{UR} \\
\midrule
Few-shot (AR)                 & --- & --- & 29.80 & 13.65 & 15.20 & 20.27 \\
Few-shot (FA)                 & --- & --- & 31.66 & 14.55 & 15.65 & 20.40 \\
Few-shot (AR+FA)              & --- & --- & 33.33 & 15.16 & 16.72 & 21.91 \\
\midrule
Adaptation (AR)               & --- & --- & 36.95 & 12.60 & 10.73 & 28.32 \\
Adaptation (FA)               & --- & --- & \textbf{38.52} & \textbf{13.14} & \textbf{11.45} & \textbf{28.79} \\
\midrule
Merge: Simple Avg             & 38.66 & 42.58 & 37.16 & 11.08 & 14.21 & 17.29 \\
Merge: Wtd.\ ($\alpha{=}0.3$) & 36.61 & 41.83 & 36.00 & 12.14 & \textbf{14.28} & 15.30 \\
Merge: Wtd.\ ($\alpha{=}0.7$) & \textbf{39.99} & 42.85 & 35.77 &  7.81 & 13.56 & 19.19 \\
Merge: TIES                   & 38.67 & \textbf{43.34} & 36.98 &  9.98 & 13.38 & 18.35 \\
Merge: DARE                   & 38.52 & 42.82 & 36.96 & 11.05 & 14.25 & 17.75 \\
\bottomrule
\end{tabular}}
\caption{CHrF++ on all target languages for few-shot transfer, adaptation ($\sim$500 sentences), and adapter merging. Stage 4 uses Llama 3.2 3B; all other stages use Gemma 2 2B. Normalised scores used where available (CAMeL for \textsc{ar}; Hazm for \textsc{fa}/\textsc{prs}). \textbf{Bold}: best per column. `---' indicates the configuration was not evaluated on that target.}
\label{tab:all-stages-results}
\end{table}

Evaluation on the held-out TICO-19 test set (2,100 sentences) closely mirrors development-set rankings, confirming that observed trends reflect genuine generalisation rather than overfitting. TIES-Merging achieves the highest score in the study on Persian (43.34 CHrF++), while adapter merging remains the strongest strategy for Pashto, with weighted averaging ($\alpha{=}0.3$) outperforming supervised adaptation (14.28 vs.\ 10.73--11.45) — confirming this as a robust effect. Dari approaches pivot quality under Simple Averaging (37.16), within 1.36 points of supervised adaptation at zero data cost, while Urdu continues to require in-language supervision, with the best merge result lagging adaptation by nearly 10 points (19.19 vs.\ 28.79). 
Sorani Kurdish 
% plateaus below deployable quality under all conditions.
remains insufficient for high-stakes clinical deployment.

\subsection{Analysis of Linguistic Similarity and Transfer}
We present in Table \ref{tab:transfer_distance} on the transfer performance compared with linguistic relationship to the pivot languages, where it shows that the transfer quality decreases as lexical and grammatical distance increase (RQ3).

\begin{table}[t]
\centering
\tiny 
\begin{tabular}{lccc}
\toprule
\textbf{Language} &
\textbf{Best CHrF++} &
\textbf{Family Relation to Persian} &
\textbf{Script} \\
\midrule
Dari   & 41.01 & Same language continuum & Shared \\
Urdu   & 28.88 & Different family        & Shared \\
Sorani & 12.97 & Iranian (distant)       & Partial \\
Pashto & 10.31 & Iranian (distant)       & Shared \\
\bottomrule
\end{tabular}
\caption{
Transfer performance compared with linguistic relationship to the
pivot languages. Languages with closer genealogical and lexical
relationships generally benefit more from cross-lingual biomedical
transfer.
}
\label{tab:transfer_distance}
\end{table}

\section{Conclusion and Future Work}
\label{sec:conc}

This study investigated whether healthcare-domain adaptation on Arabic and Persian can support neural machine translation for four underrepresented Arabic-script languages --- Dari, Pashto, Sorani Kurdish, and Urdu --- through cross-lingual transfer. We successfully fine-tuned LoRA adapters for en$\rightarrow$ar and en$\rightarrow$fa, achieving CHrF++ scores of 40.42 and 40.06 respectively under monolingual training, and up to 42.45 and 42.20 under joint training, establishing strong pivot-language baselines from which transfer could be attempted. 

Cross-lingual transfer to all four target languages was then evaluated across three complementary strategies. For Dari, the closest target language to our pivots, supervised adaptation with just 500 sentences reached near-pivot quality (CHrF++ $\approx$ 38--41), confirming that cross-lingual transfer is a practical strategy for Iranian-family languages even under severe data constraints. Zero-data adapter merging alone achieved 37.16 on the test set (within 1.36 points of supervised adaptation) making it a viable option when no target-language data is available. For Urdu, adaptation reached 28.32--28.79, demonstrating that shared script and lexical borrowing enable meaningful transfer despite its Indo-Aryan typology, though merging lagged by over 9 points, indicating that in-language supervision is necessary at this level of structural distance. 
For Pashto and Sorani Kurdish, no strategy exceeded 16 CHrF++, 
% falling well below deployable quality under all conditions, 
demonstrating the limits of cross-lingual transfer when structural distance from the pivot languages is too great. 

% Lifeng added:
The most important finding is that biomedical knowledge encoded in higher-resource Arabic and Persian adapters can be transferred to related low-resource languages through weight-space operations alone. For closely related languages such as Dari, zero-data adapter merging approaches the effectiveness of supervised adaptation, suggesting a practical path toward biomedical MT development when target-language resources are unavailable.

% \lipsum[1] % Please comment out this line and start your own content.

\section*{Limitations}
\label{sec:limitations}
% \subsection{Limitations and Validity}
The main limitation is that Pashto and Sorani Kurdish remain insufficient for high-stakes clinical deployment,
% remain below deployable quality under all strategies, 
likely due to their greater structural distance from the pivots and the domain mismatch between our general-domain adaptation data and the biomedical evaluation set. Future work should therefore focus on collecting in-domain biomedical adaptation data for these languages, identifying structurally closer pivot languages, and designing training objectives that explicitly preserve cross-lingual representations during domain specialisation.

\textbf{Limited target-language resources.}
The target languages considered in this study differ substantially in the quantity and quality of available biomedical parallel data. Although all datasets were processed using the same pipeline, differences in corpus size, annotation quality, and domain coverage may influence downstream translation performance. Consequently, some performance differences attributed to cross-lingual transfer may also partially reflect underlying dataset characteristics. Future work should evaluate transfer methods on larger and more diverse biomedical benchmarks to better isolate the effect of linguistic similarity from data availability.

\textbf{Limitations of instruction-tuned LLMs for machine translation.}
The models evaluated in this work are general-purpose instruction-tuned language models rather than systems specifically optimized for machine translation. Their outputs may therefore be affected by generation variability, hallucination, and prompt sensitivity. Furthermore, automatic metrics such as BLEU and CHRF++ may not fully capture semantic adequacy when multiple medically correct translations exist. Future work should incorporate human evaluation and semantic metrics such as COMET to provide a more comprehensive assessment of translation quality.

\textbf{Coverage of biomedical terminology.}
The biomedical corpora used in this study represent only a subset of the terminology encountered in real-world healthcare communication. As a result, the reported performance may not generalize equally across all medical subdomains, including specialized clinical, pharmaceutical, or diagnostic terminology. Although our results demonstrate effective transfer within the evaluated datasets, further validation on broader biomedical benchmarks and terminology-focused test sets is required before drawing conclusions about comprehensive biomedical translation capability.

\textbf{Confounding factors in cross-lingual transfer.}
A central hypothesis of this work is that linguistic proximity facilitates cross-lingual biomedical transfer. However, linguistic similarity is correlated with several other factors, including shared scripts, lexical overlap, corpus availability, and potential differences in pretraining exposure. Consequently, the observed transfer patterns cannot be attributed solely to language-family relationships. Future work should employ controlled comparisons and quantitative measures of linguistic distance to better disentangle these interacting effects.

\textbf{Generalizability of adapter merging.}
The effectiveness of LoRA adapter merging was evaluated using two base models and a specific biomedical adaptation setting. While the results suggest that merged adapters can transfer domain knowledge to related low-resource languages, it remains unclear whether the same behavior would generalize to larger language models, alternative adapter architectures, or non-biomedical domains. Additional experiments across broader model families and domains are needed to establish the robustness of this approach.
% This section is required by ACL-style submissions. Please revise this placeholder to discuss the main limitations of the study, including the small amount of target-language biomedical parallel data, the reliability of automatic metrics for low-resource Arabic-script languages, the lack of human medical-domain evaluation, and the risks of deploying biomedical MT systems without clinical validation.

\section*{Ethical Considerations}
\label{sec:ethics}
This work investigates biomedical machine translation for under-represented Arabic-script languages. While improved translation systems may increase access to healthcare information for speakers of low-resource languages, translation errors in clinical settings can also lead to misunderstanding, delayed treatment, or patient harm. Consequently, the models presented in this paper are intended solely for research purposes and should not be deployed for clinical decision-making without rigorous human evaluation and domain-specific validation.
Our experiments use publicly available datasets and pretrained language models. We do not collect personal health records or other sensitive patient information. Nevertheless, the biomedical corpora used in this work may inherit biases present in the original source materials, including uneven coverage of medical conditions, terminology, and language varieties. Future work should investigate fairness across dialects and demographic groups and include evaluation by professional medical translators before deployment in real-world healthcare settings.

Large language models may generate fluent but factually incorrect translations or hallucinated medical terminology. Such behaviour is particularly concerning in healthcare applications where seemingly minor translation errors can have significant consequences. Therefore, human oversight remains essential whenever these systems are used in safety-critical environments.
At the same time, improving translation quality for severely under-represented languages has the potential to increase equitable access to health information in multilingual communities that currently lack high-quality translation technology.

% The data we used for this study does not have ethical concerns.
% This section is optional under ACL guidance but recommended for biomedical MT. Please revise this placeholder to describe risks from mistranslation in clinical settings, intended non-deployment use of the current system, data privacy considerations, and the need for professional medical review before any real-world use.

\bibliography{refs}

@article{alabdullah2025advancing,
  title={Advancing dialectal Arabic to modern standard Arabic machine translation},
  author={Alabdullah, Abdullah and Han, Lifeng and Lin, Chenghua},
  journal={arXiv preprint arXiv:2507.20301},
  year={2025}
}

@misc{nllbteam2022languageleftbehindscaling_arXiv,
      title={No Language Left Behind: Scaling Human-Centered Machine Translation}, 
      author={NLLB Team and Marta R. Costa-jussà and James Cross and Onur Çelebi and Maha Elbayad and Kenneth Heafield and Kevin Heffernan and Elahe Kalbassi and Janice Lam and Daniel Licht and Jean Maillard and Anna Sun and Skyler Wang and Guillaume Wenzek and Al Youngblood and Bapi Akula and Loic Barrault and Gabriel Mejia Gonzalez and Prangthip Hansanti and John Hoffman and Semarley Jarrett and Kaushik Ram Sadagopan and Dirk Rowe and Shannon Spruit and Chau Tran and Pierre Andrews and Necip Fazil Ayan and Shruti Bhosale and Sergey Edunov and Angela Fan and Cynthia Gao and Vedanuj Goswami and Francisco Guzmán and Philipp Koehn and Alexandre Mourachko and Christophe Ropers and Safiyyah Saleem and Holger Schwenk and Jeff Wang},
      year={2022},
      eprint={2207.04672},
      archivePrefix={arXiv},
      primaryClass={cs.CL},
      url={https://arxiv.org/abs/2207.04672}, 
}

@inproceedings{han2023investigating,
    title = "Investigating Massive Multilingual Pre-Trained Machine Translation Models for Clinical Domain via Transfer Learning",
    author = "Han, Lifeng  and
      Erofeev, Gleb  and
      Sorokina, Irina  and
      Gladkoff, Serge  and
      Nenadic, Goran",
    editor = "Naumann, Tristan  and
      Ben Abacha, Asma  and
      Bethard, Steven  and
      Roberts, Kirk  and
      Rumshisky, Anna",
    booktitle = "Proceedings of the 5th Clinical Natural Language Processing Workshop",
    month = jul,
    year = "2023",
    address = "Toronto, Canada",
    publisher = "Association for Computational Linguistics",
    url = "https://aclanthology.org/2023.clinicalnlp-1.5/",
    doi = "10.18653/v1/2023.clinicalnlp-1.5",
    pages = "31--40"
}

@ARTICLE{han2024neural,
    
AUTHOR={Han, Lifeng  and Gladkoff, Serge  and Erofeev, Gleb  and Sorokina, Irina  and Galiano, Betty  and Nenadic, Goran },
           
TITLE={Neural machine translation of clinical text: an empirical investigation into multilingual pre-trained language models and transfer-learning},
          
JOURNAL={Frontiers in Digital Health},
          
VOLUME={Volume 6 - 2024},
  
YEAR={2024},
  
URL={https://www.frontiersin.org/journals/digital-health/articles/10.3389/fdgth.2024.1211564},
  
DOI={10.3389/fdgth.2024.1211564},
  
ISSN={2673-253X}}

@article{chauhan2021adableu,
author = {Shweta Chauhan and Philemon Daniel and Archita Mishra and Abhay Kumar},
title = {AdaBLEU: A Modified BLEU Score for Morphologically Rich Languages},
journal = {IETE Journal of Research},
volume = {69},
number = {8},
pages = {5112--5123},
year = {2023},
publisher = {Taylor \& Francis},
doi = {10.1080/03772063.2021.1962745},


URL = { 
    
        https://doi.org/10.1080/03772063.2021.1962745
    
    

},
eprint = { 
    
        https://doi.org/10.1080/03772063.2021.1962745
    
    

}

}

@article{bahdanau2015neural,
  title={Neural machine translation by jointly learning to align and translate},
  author={Bahdanau, Dzmitry and Cho, Kyunghyun and Bengio, Yoshua},
  journal={arXiv preprint arXiv:1409.0473},
  year={2014}
}

@inproceedings{
xu2024paradigm,
title={A Paradigm Shift in Machine Translation: Boosting Translation Performance of Large Language Models},
author={Haoran Xu and Young Jin Kim and Amr Sharaf and Hany Hassan Awadalla},
booktitle={The Twelfth International Conference on Learning Representations},
year={2024},
url={https://openreview.net/forum?id=farT6XXntP}
}

@inproceedings{zhu2024multilingual,
    title = "Multilingual Machine Translation with Large Language Models: Empirical Results and Analysis",
    author = "Zhu, Wenhao  and
      Liu, Hongyi  and
      Dong, Qingxiu  and
      Xu, Jingjing  and
      Huang, Shujian  and
      Kong, Lingpeng  and
      Chen, Jiajun  and
      Li, Lei",
    editor = "Duh, Kevin  and
      Gomez, Helena  and
      Bethard, Steven",
    booktitle = "Findings of the Association for Computational Linguistics: NAACL 2024",
    month = jun,
    year = "2024",
    address = "Mexico City, Mexico",
    publisher = "Association for Computational Linguistics",
    url = "https://aclanthology.org/2024.findings-naacl.176/",
    doi = "10.18653/v1/2024.findings-naacl.176",
    pages = "2765--2781"
}

@Article{li2026comparing,
author="Li, Andy
and Zhou, Wei
and Hoda, Rashina
and Bain, Chris
and Poon, Peter",
title="Comparing Large Language Models and Traditional Machine Translation Tools for Translating Medical Consultation Summaries: Quantitative Pilot Feasibility Study",
journal="JMIR Form Res",
year="2026",
month="Apr",
day="13",
volume="10",
pages="e85169",
issn="2561-326X",
doi="10.2196/85169",
url="https://formative.jmir.org/2026/1/e85169",
url="https://doi.org/10.2196/85169"
}

@inproceedings{
hu2022lora,
title={Lo{RA}: Low-Rank Adaptation of Large Language Models},
author={Edward J Hu and yelong shen and Phillip Wallis and Zeyuan Allen-Zhu and Yuanzhi Li and Shean Wang and Lu Wang and Weizhu Chen},
booktitle={International Conference on Learning Representations},
year={2022},
url={https://openreview.net/forum?id=nZeVKeeFYf9}
}

@inproceedings{mao-yu-2024-tuning,
    title = "Tuning {LLM}s with Contrastive Alignment Instructions for Machine Translation in Unseen, Low-resource Languages",
    author = "Mao, Zhuoyuan  and
      Yu, Yen",
    editor = "Ojha, Atul Kr.  and
      Liu, Chao-hong  and
      Vylomova, Ekaterina  and
      Pirinen, Flammie  and
      Abbott, Jade  and
      Washington, Jonathan  and
      Oco, Nathaniel  and
      Malykh, Valentin  and
      Logacheva, Varvara  and
      Zhao, Xiaobing",
    booktitle = "Proceedings of the Seventh Workshop on Technologies for Machine Translation of Low-Resource Languages (LoResMT 2024)",
    month = aug,
    year = "2024",
    address = "Bangkok, Thailand",
    publisher = "Association for Computational Linguistics",
    url = "https://aclanthology.org/2024.loresmt-1.1/",
    doi = "10.18653/v1/2024.loresmt-1.1",
    pages = "1--25"
}

@ARTICLE{xiao2025lsftl,
  author={Liang, Xiao and Jasmina Khaw, Yen-Min and Liew, Soung-Yue and Tan, Tien-Ping and Qin, Donghong},
  journal={IEEE Access}, 
  title={Toward Low-Resource Languages Machine Translation: A Language-Specific Fine-Tuning With LoRA for Specialized Large Language Models}, 
  year={2025},
  volume={13},
  number={},
  pages={46616-46626},
  doi={10.1109/ACCESS.2025.3549795}}

@inproceedings{
dettmers2023qlora,
title={{QL}o{RA}: Efficient Finetuning of Quantized {LLM}s},
author={Tim Dettmers and Artidoro Pagnoni and Ari Holtzman and Luke Zettlemoyer},
booktitle={Thirty-seventh Conference on Neural Information Processing Systems},
year={2023},
url={https://openreview.net/forum?id=OUIFPHEgJU}
}

@inproceedings{aharoni2019massively,
    title = "Massively Multilingual Neural Machine Translation",
    author = "Aharoni, Roee  and
      Johnson, Melvin  and
      Firat, Orhan",
    editor = "Burstein, Jill  and
      Doran, Christy  and
      Solorio, Thamar",
    booktitle = "Proceedings of the 2019 Conference of the North {A}merican Chapter of the Association for Computational Linguistics: Human Language Technologies, Volume 1 (Long and Short Papers)",
    month = jun,
    year = "2019",
    address = "Minneapolis, Minnesota",
    publisher = "Association for Computational Linguistics",
    url = "https://aclanthology.org/N19-1388/",
    doi = "10.18653/v1/N19-1388",
    pages = "3874--3884"
}

@inproceedings{muller2021interplay,
    title = "The interplay between language similarity and script on a novel multi-layer {A}lgerian dialect corpus",
    author = "Touileb, Samia  and
      Barnes, Jeremy",
    editor = "Zong, Chengqing  and
      Xia, Fei  and
      Li, Wenjie  and
      Navigli, Roberto",
    booktitle = "Findings of the Association for Computational Linguistics: ACL-IJCNLP 2021",
    month = aug,
    year = "2021",
    address = "Online",
    publisher = "Association for Computational Linguistics",
    url = "https://aclanthology.org/2021.findings-acl.324/",
    doi = "10.18653/v1/2021.findings-acl.324",
    pages = "3700--3712"
}

@InProceedings{wortsman2022model_soups,
  title = 	 {Model soups: averaging weights of multiple fine-tuned models improves accuracy without increasing inference time},
  author =       {Wortsman, Mitchell and Ilharco, Gabriel and Gadre, Samir Ya and Roelofs, Rebecca and Gontijo-Lopes, Raphael and Morcos, Ari S and Namkoong, Hongseok and Farhadi, Ali and Carmon, Yair and Kornblith, Simon and Schmidt, Ludwig},
  booktitle = 	 {Proceedings of the 39th International Conference on Machine Learning},
  pages = 	 {23965--23998},
  year = 	 {2022},
  editor = 	 {Chaudhuri, Kamalika and Jegelka, Stefanie and Song, Le and Szepesvari, Csaba and Niu, Gang and Sabato, Sivan},
  volume = 	 {162},
  series = 	 {Proceedings of Machine Learning Research},
  month = 	 {17--23 Jul},
  publisher =    {PMLR},
  url = 	 {https://proceedings.mlr.press/v162/wortsman22a.html}
}

@inproceedings{
ilharco2023editing,
title={Editing models with task arithmetic},
author={Gabriel Ilharco and Marco Tulio Ribeiro and Mitchell Wortsman and Ludwig Schmidt and Hannaneh Hajishirzi and Ali Farhadi},
booktitle={The Eleventh International Conference on Learning Representations },
year={2023},
url={https://openreview.net/forum?id=6t0Kwf8-jrj}
}

@inproceedings{yadav2023resolving,
 author = {Yadav, Prateek and Tam, Derek and Choshen, Leshem and Raffel, Colin and Bansal, Mohit},
 booktitle = {Advances in Neural Information Processing Systems},
 editor = {A. Oh and T. Naumann and A. Globerson and K. Saenko and M. Hardt and S. Levine},
 pages = {7093--7115},
 publisher = {Curran Associates, Inc.},
 title = {TIES-Merging: Resolving Interference When Merging Models},
 volume = {36},
 year = {2023}
}

@inproceedings{
yu2024language,
title={Language Models are Super Mario: Absorbing Abilities from Homologous Models as a Free Lunch},
author={Le Yu and Bowen Yu and Haiyang Yu and Fei Huang and Yongbin Li},
booktitle={Forty-first International Conference on Machine Learning},
year={2024},
url={https://openreview.net/forum?id=fq0NaiU8Ex}
}

@inproceedings{zhao2025adamergex,
    title = "{A}da{M}erge{X}: Cross-Lingual Transfer with Large Language Models via Adaptive Adapter Merging",
    author = "Zhao, Yiran  and
      Zhang, Wenxuan  and
      Wang, Huiming  and
      Kawaguchi, Kenji  and
      Bing, Lidong",
    editor = "Chiruzzo, Luis  and
      Ritter, Alan  and
      Wang, Lu",
    booktitle = "Proceedings of the 2025 Conference of the Nations of the Americas Chapter of the Association for Computational Linguistics: Human Language Technologies (Volume 1: Long Papers)",
    month = apr,
    year = "2025",
    address = "Albuquerque, New Mexico",
    publisher = "Association for Computational Linguistics",
    url = "https://aclanthology.org/2025.naacl-long.493/",
    doi = "10.18653/v1/2025.naacl-long.493",
    pages = "9785--9800",
    ISBN = "979-8-89176-189-6"
}

@inproceedings{bandarkar2025unreasonable,
    title = "The Unreasonable Effectiveness of Model Merging for Cross-Lingual Transfer in {LLM}s",
    author = "Bandarkar, Lucas  and
      Peng, Nanyun",
    editor = "Adelani, David Ifeoluwa  and
      Arnett, Catherine  and
      Ataman, Duygu  and
      Chang, Tyler A.  and
      Gonen, Hila  and
      Raja, Rahul  and
      Schmidt, Fabian  and
      Stap, David  and
      Wang, Jiayi",
    booktitle = "Proceedings of the 5th Workshop on Multilingual Representation Learning (MRL 2025)",
    month = nov,
    year = "2025",
    address = "Suzhuo, China",
    publisher = "Association for Computational Linguistics",
    url = "https://aclanthology.org/2025.mrl-main.10/",
    doi = "10.18653/v1/2025.mrl-main.10",
    pages = "131--148",
    ISBN = "979-8-89176-345-6"
}

@inproceedings{chronopoulou2023language,
    title = "Language-Family Adapters for Low-Resource Multilingual Neural Machine Translation",
    author = "Chronopoulou, Alexandra  and
      Stojanovski, Dario  and
      Fraser, Alexander",
    editor = "Ojha, Atul Kr.  and
      Liu, Chao-hong  and
      Vylomova, Ekaterina  and
      Pirinen, Flammie  and
      Abbott, Jade  and
      Washington, Jonathan  and
      Oco, Nathaniel  and
      Malykh, Valentin  and
      Logacheva, Varvara  and
      Zhao, Xiaobing",
    booktitle = "Proceedings of the Sixth Workshop on Technologies for Machine Translation of Low-Resource Languages (LoResMT 2023)",
    month = may,
    year = "2023",
    address = "Dubrovnik, Croatia",
    publisher = "Association for Computational Linguistics",
    url = "https://aclanthology.org/2023.loresmt-1.5/",
    doi = "10.18653/v1/2023.loresmt-1.5",
    pages = "59--72"
}

@inproceedings{tao2024unlocking,
    title = "Unlocking the Potential of Model Merging for Low-Resource Languages",
    author = "Tao, Mingxu  and
      Zhang, Chen  and
      Huang, Quzhe  and
      Ma, Tianyao  and
      Huang, Songfang  and
      Zhao, Dongyan  and
      Feng, Yansong",
    editor = "Al-Onaizan, Yaser  and
      Bansal, Mohit  and
      Chen, Yun-Nung",
    booktitle = "Findings of the Association for Computational Linguistics: EMNLP 2024",
    month = nov,
    year = "2024",
    address = "Miami, Florida, USA",
    publisher = "Association for Computational Linguistics",
    url = "https://aclanthology.org/2024.findings-emnlp.508/",
    doi = "10.18653/v1/2024.findings-emnlp.508",
    pages = "8705--8720"
}

@article{zappatore2024adopting,
  title={Adopting machine translation in the healthcare sector: A methodological multi-criteria review},
  author={Zappatore, Marco and Ruggieri, Gilda},
  journal={Computer Speech \& Language},
  volume={84},
  pages={101582},
  year={2024},
  publisher={Elsevier}
}

@inproceedings{neves2024findings,
    title = "Findings of the {WMT} 2024 Biomedical Translation Shared Task: Test Sets on Abstract Level",
    author = "Neves, Mariana  and
      Grozea, Cristian  and
      Thomas, Philippe  and
      Roller, Roland  and
      Bawden, Rachel  and
      N{\'e}v{\'e}ol, Aur{\'e}lie  and
      Castle, Steffen  and
      Bonato, Vanessa  and
      Di Nunzio, Giorgio Maria  and
      Vezzani, Federica  and
      Vicente Navarro, Maika  and
      Yeganova, Lana  and
      Jimeno Yepes, Antonio",
    editor = "Haddow, Barry  and
      Kocmi, Tom  and
      Koehn, Philipp  and
      Monz, Christof",
    booktitle = "Proceedings of the Ninth Conference on Machine Translation",
    month = nov,
    year = "2024",
    address = "Miami, Florida, USA",
    publisher = "Association for Computational Linguistics",
    url = "https://aclanthology.org/2024.wmt-1.6/",
    doi = "10.18653/v1/2024.wmt-1.6",
    pages = "124--138"
}

@inproceedings{mehandru2023physician,
    title = "Physician Detection of Clinical Harm in Machine Translation: Quality Estimation Aids in Reliance and Backtranslation Identifies Critical Errors",
    author = "Mehandru, Nikita  and
      Agrawal, Sweta  and
      Xiao, Yimin  and
      Gao, Ge  and
      Khoong, Elaine  and
      Carpuat, Marine  and
      Salehi, Niloufar",
    editor = "Bouamor, Houda  and
      Pino, Juan  and
      Bali, Kalika",
    booktitle = "Proceedings of the 2023 Conference on Empirical Methods in Natural Language Processing",
    month = dec,
    year = "2023",
    address = "Singapore",
    publisher = "Association for Computational Linguistics",
    url = "https://aclanthology.org/2023.emnlp-main.712/",
    doi = "10.18653/v1/2023.emnlp-main.712",
    pages = "11633--11647"
}

@inproceedings{lakew2018comparison,
    title = "A Comparison of Transformer and Recurrent Neural Networks on Multilingual Neural Machine Translation",
    author = "Lakew, Surafel Melaku  and
      Cettolo, Mauro  and
      Federico, Marcello",
    editor = "Bender, Emily M.  and
      Derczynski, Leon  and
      Isabelle, Pierre",
    booktitle = "Proceedings of the 27th International Conference on Computational Linguistics",
    month = aug,
    year = "2018",
    address = "Santa Fe, New Mexico, USA",
    publisher = "Association for Computational Linguistics",
    url = "https://aclanthology.org/C18-1054/",
    pages = "641--652"
}

@article{johnson2017googles,
    title = "{G}oogle{'}s Multilingual Neural Machine Translation System: Enabling Zero-Shot Translation",
    author = "Johnson, Melvin  and
      Schuster, Mike  and
      Le, Quoc V.  and
      Krikun, Maxim  and
      Wu, Yonghui  and
      Chen, Zhifeng  and
      Thorat, Nikhil  and
      Vi{\'e}gas, Fernanda  and
      Wattenberg, Martin  and
      Corrado, Greg  and
      Hughes, Macduff  and
      Dean, Jeffrey",
    editor = "Lee, Lillian  and
      Johnson, Mark  and
      Toutanova, Kristina",
    journal = "Transactions of the Association for Computational Linguistics",
    volume = "5",
    year = "2017",
    address = "Cambridge, MA",
    publisher = "MIT Press",
    url = "https://aclanthology.org/Q17-1024/",
    doi = "10.1162/tacl_a_00065",
    pages = "339--351"
}

@inproceedings{esalati2024esposito,
    title = "Esposito: An {E}nglish-{P}ersian Scientific Parallel Corpus for Machine Translation",
    author = "Esalati, Mersad  and
      Dousti, Mohammad Javad  and
      Faili, Heshaam",
    editor = "Calzolari, Nicoletta  and
      Kan, Min-Yen  and
      Hoste, Veronique  and
      Lenci, Alessandro  and
      Sakti, Sakriani  and
      Xue, Nianwen",
    booktitle = "Proceedings of the 2024 Joint International Conference on Computational Linguistics, Language Resources and Evaluation (LREC-COLING 2024)",
    month = may,
    year = "2024",
    address = "Torino, Italia",
    publisher = "ELRA and ICCL",
    url = "https://aclanthology.org/2024.lrec-main.557/",
    pages = "6299--6308"
}

@software{han2023unsloth,
  author = {Daniel Han, Michael Han and Unsloth team},
  title = {Unsloth},
  url = {https://github.com/unslothai/unsloth},
  year = {2023}
}

@article{peach2023,
author = {Al-Sabbagh, Rania},
title = {PEACH: a sentence-aligned Parallel English–Arabic Corpus for Healthcare},
journal = {Corpora},
volume = {19},
number = {3},
pages = {395-410},
year = {2024},
doi = {10.3366/cor.2024.0320},

URL = { 
    
        https://doi.org/10.3366/cor.2024.0320
    
    

},
eprint = { 
    
        https://doi.org/10.3366/cor.2024.0320
    
    

}
}

@inproceedings{tico,
    title = "{TICO}-19: the Translation Initiative for {CO}vid-19",
    author = {Anastasopoulos, Antonios  and
      Cattelan, Alessandro  and
      Dou, Zi-Yi  and
      Federico, Marcello  and
      Federmann, Christian  and
      Genzel, Dmitriy  and
      Guzm{\'a}n, Francisco  and
      Hu, Junjie  and
      Hughes, Macduff  and
      Koehn, Philipp  and
      Lazar, Rosie  and
      Lewis, Will  and
      Neubig, Graham  and
      Niu, Mengmeng  and
      {\"O}ktem, Alp  and
      Paquin, Eric  and
      Tang, Grace  and
      Tur, Sylwia},
    editor = "Verspoor, Karin  and
      Cohen, Kevin Bretonnel  and
      Conway, Michael  and
      de Bruijn, Berry  and
      Dredze, Mark  and
      Mihalcea, Rada  and
      Wallace, Byron",
    booktitle = "Proceedings of the 1st Workshop on {NLP} for {COVID}-19 (Part 2) at {EMNLP} 2020",
    month = dec,
    year = "2020",
    address = "Online",
    publisher = "Association for Computational Linguistics",
    url = "https://aclanthology.org/2020.nlpcovid19-2.5/",
    doi = "10.18653/v1/2020.nlpcovid19-2.5"
}

@inproceedings{merx2025openwho,
    title = "{O}pen{WHO}: A Document-Level Parallel Corpus for Health Translation in Low-Resource Languages",
    author = "Merx, Raphael  and
      Suominen, Hanna  and
      Cohn, Trevor  and
      Vylomova, Ekaterina",
    editor = "Haddow, Barry  and
      Kocmi, Tom  and
      Koehn, Philipp  and
      Monz, Christof",
    booktitle = "Proceedings of the Tenth Conference on Machine Translation",
    month = nov,
    year = "2025",
    address = "Suzhou, China",
    publisher = "Association for Computational Linguistics",
    url = "https://aclanthology.org/2025.wmt-1.8/",
    doi = "10.18653/v1/2025.wmt-1.8",
    pages = "142--160",
    ISBN = "979-8-89176-341-8"
}

@inproceedings{brown2020gpt3,
 author = {Brown, Tom and Mann, Benjamin and Ryder, Nick and Subbiah, Melanie and Kaplan, Jared D and Dhariwal, Prafulla and Neelakantan, Arvind and Shyam, Pranav and Sastry, Girish and Askell, Amanda and Agarwal, Sandhini and Herbert-Voss, Ariel and Krueger, Gretchen and Henighan, Tom and Child, Rewon and Ramesh, Aditya and Ziegler, Daniel and Wu, Jeffrey and Winter, Clemens and Hesse, Chris and Chen, Mark and Sigler, Eric and Litwin, Mateusz and Gray, Scott and Chess, Benjamin and Clark, Jack and Berner, Christopher and McCandlish, Sam and Radford, Alec and Sutskever, Ilya and Amodei, Dario},
 booktitle = {Advances in Neural Information Processing Systems},
 editor = {H. Larochelle and M. Ranzato and R. Hadsell and M.F. Balcan and H. Lin},
 pages = {1877--1901},
 publisher = {Curran Associates, Inc.},
 title = {Language Models are Few-Shot Learners},
 volume = {33},
 year = {2020}
}

@inproceedings{rei-etal-2020-comet,
    title = "{COMET}: A Neural Framework for {MT} Evaluation",
    author = "Rei, Ricardo  and
      Stewart, Craig  and
      Farinha, Ana C  and
      Lavie, Alon",
    editor = "Webber, Bonnie  and
      Cohn, Trevor  and
      He, Yulan  and
      Liu, Yang",
    booktitle = "Proceedings of the 2020 Conference on Empirical Methods in Natural Language Processing (EMNLP)",
    month = nov,
    year = "2020",
    address = "Online",
    publisher = "Association for Computational Linguistics",
    url = "https://aclanthology.org/2020.emnlp-main.213/",
    doi = "10.18653/v1/2020.emnlp-main.213",
    pages = "2685--2702"
}

@inproceedings{sutskever2014sequence,
  title={Sequence to Sequence Learning with Neural Networks},
  author={Sutskever, Ilya and Vinyals, Oriol and Le, Quoc V},
  booktitle={Advances in Neural Information Processing Systems},
  volume={27},
  pages={3104--3112},
  year={2014}
}

@inproceedings{conneau2020unsupervised,
    title = "Unsupervised Cross-lingual Representation Learning at Scale",
    author = "Conneau, Alexis  and
      Khandelwal, Kartikay  and
      Goyal, Naman  and
      Chaudhary, Vishrav  and
      Wenzek, Guillaume  and
      Guzm{\'a}n, Francisco  and
      Grave, Edouard  and
      Ott, Myle  and
      Zettlemoyer, Luke  and
      Stoyanov, Veselin",
    editor = "Jurafsky, Dan  and
      Chai, Joyce  and
      Schluter, Natalie  and
      Tetreault, Joel",
    booktitle = "Proceedings of the 58th Annual Meeting of the Association for Computational Linguistics",
    month = jul,
    year = "2020",
    address = "Online",
    publisher = "Association for Computational Linguistics",
    url = "https://aclanthology.org/2020.acl-main.747/",
    doi = "10.18653/v1/2020.acl-main.747",
    pages = "8440--8451"
}

@article{pan2010survey,
author = {Pan, Sinno Jialin and Yang, Qiang},
title = {A Survey on Transfer Learning},
year = {2010},
issue_date = {October 2010},
publisher = {IEEE Educational Activities Department},
address = {USA},
volume = {22},
number = {10},
issn = {1041-4347},
url = {https://doi.org/10.1109/TKDE.2009.191},
doi = {10.1109/TKDE.2009.191},
journal = {IEEE Trans. on Knowl. and Data Eng.},
month = oct,
pages = {1345–1359},
numpages = {15}
}

@InProceedings{garcia2023unreasonable,
  title = 	 {The Unreasonable Effectiveness of Few-shot Learning for Machine Translation},
  author =       {Garcia, Xavier and Bansal, Yamini and Cherry, Colin and Foster, George and Krikun, Maxim and Johnson, Melvin and Firat, Orhan},
  booktitle = 	 {Proceedings of the 40th International Conference on Machine Learning},
  pages = 	 {10867--10878},
  year = 	 {2023},
  editor = 	 {Krause, Andreas and Brunskill, Emma and Cho, Kyunghyun and Engelhardt, Barbara and Sabato, Sivan and Scarlett, Jonathan},
  volume = 	 {202},
  series = 	 {Proceedings of Machine Learning Research},
  month = 	 {23--29 Jul},
  publisher =    {PMLR},
  url = 	 {https://proceedings.mlr.press/v202/garcia23a.html}
}

@inproceedings{10.3115/1073083.1073135,
author = {Papineni, Kishore and Roukos, Salim and Ward, Todd and Zhu, Wei-Jing},
title = {BLEU: a method for automatic evaluation of machine translation},
year = {2002},
publisher = {Association for Computational Linguistics},
address = {USA},
url = {https://doi.org/10.3115/1073083.1073135},
doi = {10.3115/1073083.1073135},
booktitle = {Proceedings of the 40th Annual Meeting on Association for Computational Linguistics},
pages = {311–318},
numpages = {8},
location = {Philadelphia, Pennsylvania},
series = {ACL '02}
}

@inproceedings{popovic-2015-chrf,
    title = "chr{F}: character n-gram {F}-score for automatic {MT} evaluation",
    author = "Popovi{\'c}, Maja",
    editor = "Bojar, Ond{\v{r}}ej  and
      Chatterjee, Rajan  and
      Federmann, Christian  and
      Haddow, Barry  and
      Hokamp, Chris  and
      Huck, Matthias  and
      Logacheva, Varvara  and
      Pecina, Pavel",
    booktitle = "Proceedings of the Tenth Workshop on Statistical Machine Translation",
    month = sep,
    year = "2015",
    address = "Lisbon, Portugal",
    publisher = "Association for Computational Linguistics",
    url = "https://aclanthology.org/W15-3049/",
    doi = "10.18653/v1/W15-3049",
    pages = "392--395"
}

@inproceedings{popovic-2017-chrf,
    title = "chr{F}++: words helping character n-grams",
    author = "Popovi{\'c}, Maja",
    editor = "Bojar, Ond{\v{r}}ej  and
      Buck, Christian  and
      Chatterjee, Rajen  and
      Federmann, Christian  and
      Graham, Yvette  and
      Haddow, Barry  and
      Huck, Matthias  and
      Yepes, Antonio Jimeno  and
      Koehn, Philipp  and
      Kreutzer, Julia",
    booktitle = "Proceedings of the Second Conference on Machine Translation",
    month = sep,
    year = "2017",
    address = "Copenhagen, Denmark",
    publisher = "Association for Computational Linguistics",
    url = "https://aclanthology.org/W17-4770/",
    doi = "10.18653/v1/W17-4770",
    pages = "612--618"
}

@inproceedings{obeid-etal-2020-camel,
    title = "{CAM}e{L} Tools: An Open Source Python Toolkit for {A}rabic Natural Language Processing",
    author = "Obeid, Ossama  and
      Zalmout, Nasser  and
      Khalifa, Salam  and
      Taji, Dima  and
      Oudah, Mai  and
      Alhafni, Bashar  and
      Inoue, Go  and
      Eryani, Fadhl  and
      Erdmann, Alexander  and
      Habash, Nizar",
    editor = "Calzolari, Nicoletta  and
      B{\'e}chet, Fr{\'e}d{\'e}ric  and
      Blache, Philippe  and
      Choukri, Khalid  and
      Cieri, Christopher  and
      Declerck, Thierry  and
      Goggi, Sara  and
      Isahara, Hitoshi  and
      Maegaard, Bente  and
      Mariani, Joseph  and
      Mazo, H{\'e}l{\`e}ne  and
      Moreno, Asuncion  and
      Odijk, Jan  and
      Piperidis, Stelios",
    booktitle = "Proceedings of the Twelfth Language Resources and Evaluation Conference",
    month = may,
    year = "2020",
    address = "Marseille, France",
    publisher = "European Language Resources Association",
    url = "https://aclanthology.org/2020.lrec-1.868/",
    pages = "7022--7032",
    language = "eng",
    ISBN = "979-10-95546-34-4"
}

@software{hazm,
  author    = {Roshan},
  title     = {Hazm - Persian NLP Toolkit},
  url       = {https://github.com/roshan-research/hazm},
}

@inproceedings{abouzahir-etal-2026-cross,
    title = "Cross-Lingual Empirical Evaluation of Large Language Models for {A}rabic Medical Tasks",
    author = "Abouzahir, Chaimae  and
      Ma, Congbo  and
      Habash, Nizar  and
      Shamout, Farah E.",
    editor = {Danilova, Vera  and
      Kurfal{\i}, Murathan  and
      S{\"o}derfeldt, Ylva  and
      Reed, Julia  and
      Burchell, Andrew},
    booktitle = "Proceedings of the 1st Workshop on Linguistic Analysis for Health ({H}ea{L}ing 2026)",
    month = mar,
    year = "2026",
    address = "Rabat, Morocco",
    publisher = "Association for Computational Linguistics",
    url = "https://aclanthology.org/2026.healing-1.13/",
    doi = "10.18653/v1/2026.healing-1.13",
    pages = "158--171",
    ISBN = "979-8-89176-367-8"
}

@inproceedings{nigatu-etal-2025-viability,
    title = "Viability of Machine Translation for Healthcare in Low-Resourced Languages",
    author = "Nigatu, Hellina Hailu  and
      Mehandru, Nikita  and
      Abadi, Negasi Haile  and
      Gebremeskel, Blen  and
      Alaa, Ahmed  and
      Choudhury, Monojit",
    editor = "Christodoulopoulos, Christos  and
      Chakraborty, Tanmoy  and
      Rose, Carolyn  and
      Peng, Violet",
    booktitle = "Proceedings of the 2025 Conference on Empirical Methods in Natural Language Processing",
    month = nov,
    year = "2025",
    address = "Suzhou, China",
    publisher = "Association for Computational Linguistics",
    url = "https://aclanthology.org/2025.emnlp-main.535/",
    doi = "10.18653/v1/2025.emnlp-main.535",
    pages = "10584--10598",
    ISBN = "979-8-89176-332-6"
}

@inproceedings{bawden-etal-2019-findings,
    title = "Findings of the {WMT} 2019 Biomedical Translation Shared Task: Evaluation for {MEDLINE} Abstracts and Biomedical Terminologies",
    author = "Bawden, Rachel  and
      Bretonnel Cohen, Kevin  and
      Grozea, Cristian  and
      Jimeno Yepes, Antonio  and
      Kittner, Madeleine  and
      Krallinger, Martin  and
      Mah, Nancy  and
      Neveol, Aurelie  and
      Neves, Mariana  and
      Soares, Felipe  and
      Siu, Amy  and
      Verspoor, Karin  and
      Vicente Navarro, Maika",
    editor = "Bojar, Ond{\v{r}}ej  and
      Chatterjee, Rajen  and
      Federmann, Christian  and
      Fishel, Mark  and
      Graham, Yvette  and
      Haddow, Barry  and
      Huck, Matthias  and
      Yepes, Antonio Jimeno  and
      Koehn, Philipp  and
      Martins, Andr{\'e}  and
      Monz, Christof  and
      Negri, Matteo  and
      N{\'e}v{\'e}ol, Aur{\'e}lie  and
      Neves, Mariana  and
      Post, Matt  and
      Turchi, Marco  and
      Verspoor, Karin",
    booktitle = "Proceedings of the Fourth Conference on Machine Translation (Volume 3: Shared Task Papers, Day 2)",
    month = aug,
    year = "2019",
    address = "Florence, Italy",
    publisher = "Association for Computational Linguistics",
    url = "https://aclanthology.org/W19-5403/",
    doi = "10.18653/v1/W19-5403",
    pages = "29--53"
}

@misc{talwar2025pivotlanguagelowresourcemachine,
      title={Pivot Language for Low-Resource Machine Translation}, 
      author={Abhimanyu Talwar and Julien Laasri},
      year={2025},
      eprint={2505.14553},
      archivePrefix={arXiv},
      primaryClass={cs.CL},
      url={https://arxiv.org/abs/2505.14553}, 
}

@inproceedings{wu-etal-2024-far,
    title = "How Far can 100 Samples Go? Unlocking Zero-Shot Translation with Tiny Multi-Parallel Data",
    author = "Wu, Di  and
      Tan, Shaomu  and
      Meng, Yan  and
      Stap, David  and
      Monz, Christof",
    editor = "Ku, Lun-Wei  and
      Martins, Andre  and
      Srikumar, Vivek",
    booktitle = "Findings of the Association for Computational Linguistics: ACL 2024",
    month = aug,
    year = "2024",
    address = "Bangkok, Thailand",
    publisher = "Association for Computational Linguistics",
    url = "https://aclanthology.org/2024.findings-acl.896/",
    doi = "10.18653/v1/2024.findings-acl.896",
    pages = "15092--15108"
}

@inproceedings{
loshchilov2019decoupledweightdecayregularization,
title={Decoupled Weight Decay Regularization},
author={Ilya Loshchilov and Frank Hutter},
booktitle={International Conference on Learning Representations},
year={2019},
url={https://openreview.net/forum?id=Bkg6RiCqY7},
}

@misc{wolf2020huggingfacestransformersstateoftheartnatural,
      title={HuggingFace's Transformers: State-of-the-art Natural Language Processing}, 
      author={Thomas Wolf and Lysandre Debut and Victor Sanh and Julien Chaumond and Clement Delangue and Anthony Moi and Pierric Cistac and Tim Rault and Rémi Louf and Morgan Funtowicz and Joe Davison and Sam Shleifer and Patrick von Platen and Clara Ma and Yacine Jernite and Julien Plu and Canwen Xu and Teven Le Scao and Sylvain Gugger and Mariama Drame and Quentin Lhoest and Alexander M. Rush},
      year={2020},
      eprint={1910.03771},
      archivePrefix={arXiv},
      primaryClass={cs.CL},
      url={https://arxiv.org/abs/1910.03771}, 
}

@InProceedings{houlsby2019parameterefficienttransferlearningnlp,
  title = 	 {Parameter-Efficient Transfer Learning for {NLP}},
  author =       {Houlsby, Neil and Giurgiu, Andrei and Jastrzebski, Stanislaw and Morrone, Bruna and De Laroussilhe, Quentin and Gesmundo, Andrea and Attariyan, Mona and Gelly, Sylvain},
  booktitle = 	 {Proceedings of the 36th International Conference on Machine Learning},
  pages = 	 {2790--2799},
  year = 	 {2019},
  editor = 	 {Chaudhuri, Kamalika and Salakhutdinov, Ruslan},
  volume = 	 {97},
  series = 	 {Proceedings of Machine Learning Research},
  month = 	 {09--15 Jun},
  publisher =    {PMLR},
  url = 	 {https://proceedings.mlr.press/v97/houlsby19a.html}
}

@software{vonwerra2020trl,
author = {von Werra, Leandro and Belkada, Younes and Tunstall, Lewis and Beeching, Edward and Thrush, Tristan and Lambert, Nathan and Huang, Shengyi and Rasul, Kashif and Gallouédec, Quentin},
license = {Apache-2.0},
month = mar,
title = {{TRL: Transformers Reinforcement Learning}},
url = {https://github.com/huggingface/trl},
version = {1.8},
year = {2020}
}

@INPROCEEDINGS{inproceedings,
   AUTHOR = {Raiomond Doctor and Alexander Gutkin and Cibu Johny and Brian Roark and Richard Sproat},
   EDITOR = {Haralambous, Yannis},
   TITLE = {{Graphemic Normalization of the Perso-Arabic Script}},
   BOOKTITLE = {{Proceedings of Grapholinguistics in the 21st Century, 2022}},
   SERIES = {{Grapholinguistics and Its Applications}},
   VOLUME = {9},
   PUBLISHER = {Fluxus Editions},
   ADDRESS = {Brest},
   YEAR = {2022},
   PAGES = {315--376},
   DOI = {https://doi.org/10.36824/2022-graf-gutk},
}

@article{al_shamsi_implications_2020,
  title={Implications of language barriers for healthcare: a systematic review},
  author={Al Shamsi, Hilal and Almutairi, Abdullah G and Al Mashrafi, Sulaiman and Al Kalbani, Talib},
  journal={Oman medical journal},
  volume={35},
  number={2},
  pages={e122},
  year={2020}
}

@inproceedings{koehn-etal-2007-moses,
    title = "{M}oses: Open Source Toolkit for Statistical Machine Translation",
    author = "Koehn, Philipp  and
      Hoang, Hieu  and
      Birch, Alexandra  and
      Callison-Burch, Chris  and
      Federico, Marcello  and
      Bertoldi, Nicola  and
      Cowan, Brooke  and
      Shen, Wade  and
      Moran, Christine  and
      Zens, Richard  and
      Dyer, Chris  and
      Bojar, Ond{\v{r}}ej  and
      Constantin, Alexandra  and
      Herbst, Evan",
    editor = "Ananiadou, Sophia",
    booktitle = "Proceedings of the 45th Annual Meeting of the Association for Computational Linguistics Companion Volume Proceedings of the Demo and Poster Sessions",
    month = jun,
    year = "2007",
    address = "Prague, Czech Republic",
    publisher = "Association for Computational Linguistics",
    url = "https://aclanthology.org/P07-2045/",
    pages = "177--180"
}

\newpage

\appendix

\section{Model Merging Details}
\label{appendix:model-merge-details}

\subsection{Weighted Averaging}

The simplest merging strategy is weighted averaging, which interpolates between two task vectors via a blending coefficient $\alpha \in [0, 1]$:
\begin{equation}
    \theta_{\text{merged}} = \theta_0 + \alpha \tau_1 + (1 - \alpha) \tau_2.
\end{equation}
Simple averaging is the special case $\alpha = 0.5$, treating both adapters as equally informative. The general weighted form is more appropriate for our cross-lingual setting, where the two pivot languages are not equally related to each target language. By biasing toward the Persian adapter ($\alpha \to 0$) or the Arabic adapter ($\alpha \to 1$), we can exploit known linguistic proximity without any additional training. For Iranian-family targets such as Dari and Sorani Kurdish, Persian-biased weighting is a principled choice; for Urdu, where Arabic script influence is stronger, Arabic-biased weighting is preferred.

This approach is well-motivated when both adapters share the same frozen base model and LoRA initialisation, as their task vectors lie in a comparable parameter space. Arabic and Persian share script, morphological patterns, and biomedical vocabulary, making it plausible that their adapter updates encode compatible representations. When this compatibility holds, the interpolated midpoint remains a low-loss solution for both languages \cite{wortsman2022model_soups}.

 \subsection{TIES-Merging}                                                                                                                                                        
   
  % Weighted averaging can fail when two adapters assign opposite signs to the same                                                                                                     
  % parameter, causing partial cancellation. In our setting, Arabic and Persian morphological
  % divergence can produce precisely such opposing updates. TIES-Merging~\cite{yadav2023resolving}                                                                                      
  % resolves this through three steps: it first trims each task vector by zeroing out
  % low-magnitude parameters — retaining only the top-$k$ fraction by absolute magnitude —                                                                                              
  % to suppress noise; it then elects a dominant sign per parameter position based on
  % aggregate signed mass across adapters; finally, it averages only those adapter values                                                                                               
  % that agree with the elected sign, excluding conflicting updates entirely. This prevents
  % the destructive cancellation that arises when Arabic and Persian encode the same                                                                                                    
  % biomedical concept through morphologically divergent parameter updates.
 Weighted averaging can fail when adapters assign opposite signs to the same parameter,                                                                                              
  causing destructive cancellation — a concrete risk given Arabic--Persian morphological                                                                                              
  divergence. TIES-Merging~\cite{yadav2023resolving} mitigates this via three operations                                                                                              
  on each task vector $\tau_i = \theta_i - \theta_0$: \textit{trim}, which zeros all                                                                                                  
  parameters outside the top-$k$ absolute-magnitude mass, discarding low-signal noise;                                                                                                
  \textit{elect}, which determines a dominant sign per position from the aggregate signed                                                                                             
  mass $\sum_i \tilde{\tau}_i[j]$; and \textit{disjoint merge}, which averages only                                                                                                   
  the adapter values consistent with the elected sign, fully excluding conflicting                                                                                                    
  updates. The result preserves the dominant task-specific direction at each parameter                                                                                                
  position rather than collapsing opposing updates into a attenuated compromise.                                                                                                      
           
\subsection{DARE}

Drop And REscale (DARE) \cite{yu2024language} addresses interference from a complementary angle. Rather than resolving conflicts deterministically, it reduces the probability that conflicting updates co-occur at the same position at all. This is motivated by the observation that many fine-tuned parameter updates are redundant: in our setting, where Arabic and Persian share substantial biomedical vocabulary, both adapters may reinforce the same domain signal through overlapping parameters. By independently subsampling each adapter before merging, DARE reduces unnecessary overlap without requiring prior knowledge of which parameters conflict.

Formally, for each adapter $i$ and parameter $j$, a Bernoulli mask $m_{ij} \sim \text{Bernoulli}(1 - p)$ is sampled, where $p$ is the drop rate. The masked adapter is rescaled to preserve the expected magnitude of the update:

\begin{equation}
    \tilde{\tau}_i[j] = \frac{m_{ij}}{1 - p} \cdot \tau_i[j].
\end{equation}

The rescaling factor $1/(1-p)$ ensures $\mathbb{E}[\tilde{\tau}_i[j]] = \tau_i[j]$. The merged adapter is then:

\begin{equation}
    \theta_{\text{merged}} = \frac{1}{2}(\tilde{\tau}_1 + \tilde{\tau}_2) + \theta_0.
\end{equation}

At $p = 0.3$, the probability that both adapters retain the same parameter is $(1-p)^2 = 0.49$, halving the expected number of conflicting positions relative to standard averaging. Unlike TIES, which excludes conflicting parameters entirely, DARE reduces interference stochastically — a softer form of conflict resolution that retains more of each adapter's signal at the cost of some residual noise.

\section{Experimental Investigation Details}
\label{sec:exps}

% We investigate our research question (\S\ref{sec:intro}) using a six-stage experimental pipeline that progressively builds toward our cross-lingual transfer goal. We first describe the configurations used across experiments (\S\ref{sec:exp_setup}), followed by model selection, and monolingual and joint multilingual fine-tuning (\S\ref{sec:pivot_training}). We then introduce three complementary transfer strategies: few-shot in-context learning (\S\ref{sec:fewshot}), adaptation fine-tuning (\S\ref{sec:adapt_ft}), and zero-data adapter merging (\S\ref{sec:adapter_merge}). For each experiment, we explain its motivation and relevance to the research objective, followed by a technical description of the experimental procedure.

\subsection{Experimental Setup}
\label{sec:exp_setup}

All models are loaded with 4-bit NormalFloat \cite{dettmers2023qlora} quantisation via \texttt{bitsandbytes}. Experiments are run on NVIDIA L4 and A40 GPUs, enabling bfloat16 mixed precision for both training and inference. Models supported by Unsloth (Gemma 2 2B, Llama 3.2 3B, and Qwen 2.5 1.5B) are loaded via Unsloth's \texttt{FastLanguageModel}, while incompatible architectures (BLOOM 1.7B, mGPT 1.3B, and XGLM 1.7B) are loaded via HuggingFace Transformers and PEFT with the same quantisation setup.

Parameter-efficient fine-tuning is applied using LoRA~\cite{hu2022lora}, with rank $r = 16$, scaling factor $\alpha = 16$ (i.e.\ $\alpha/r = 1.0$), and no dropout, applied to all seven projection matrices: \texttt{q\_proj}, \texttt{k\_proj}, \texttt{v\_proj}, \texttt{o\_proj}, \texttt{gate\_proj}, \texttt{up\_proj}, and \texttt{down\_proj}.

Optimisation is performed using AdamW (8-bit)~\cite{loshchilov2019decoupledweightdecayregularization} with a learning rate of $1\times10^{-4}$, linear decay, 50 warmup steps, and weight decay of 0.01. The effective batch size is 32, and the maximum sequence length is 256 tokens. Sequence packing is disabled to prevent cross-contamination between training examples. All runs use 1 epoch to avoid memorisation. A fixed random seed of 3407 is used throughout. The final checkpoint is used for all evaluations.

All models use the same Alpaca-style instruction template (see Appendix), ensuring differences in translation quality arise from model or training configuration rather than prompting. We use greedy decoding throughout to ensure that performance differences across experiments reflect model and training configuration only, not decoding hyperparameters. All fine-tuning is implemented using HuggingFace Transformers \cite{wolf2020huggingfacestransformersstateoftheartnatural}, PEFT \cite{houlsby2019parameterefficienttransferlearningnlp}, and TRL's \texttt{SFTTrainer} \cite{vonwerra2020trl}. Unless otherwise specified, all experiments in this section follow the above-mentioned configurations. More information about our experimental setup can be found in Appendix~A.

\subsection{Pivot Language Adapter Training}
\label{sec:pivot_training}

Before fine-tuning, we identify the most suitable base model by evaluating six small (1B--3B) open-source decoder-only LLMs: Gemma 2 2B, Llama 3.2 3B, Qwen 2.5 1.5B, BLOOM 1.7B, mGPT 1.3B, and XGLM 1.7B. Each is evaluated zero-shot across all seven translation directions and fine-tuned on 500 sentence pairs from our en$\rightarrow$ar and en$\rightarrow$fa training sets (\S\ref{trainingdata}), using a reduced learning rate ($5 \times 10^{-5}$) and warmup (5 steps). This identifies which models exhibit strong multilingual capability and adapt efficiently with minimal data — both prerequisites for cross-lingual transfer.

We then fine-tune LoRA adapters for the two best-performing models on en$\rightarrow$ar and en$\rightarrow$fa using the full 25,000-pair training sets. These monolingual adapters serve two roles: establishing strong pivot-language baselines, and acting as inputs to all subsequent transfer experiments (\S\S~\ref{sec:fewshot}--\ref{sec:adapter_merge}).

Finally, we train a joint AR+FA adapter by concatenating, shuffling, and formatting both training sets with a language-conditioned prompt. Only Gemma 2 2B and Llama 3.2 3B are used. The LoRA rank is increased to 32 ($\alpha = 64$) to provide sufficient capacity for multilingual encoding. This joint adapter serves as an upper bound for our zero-data adapter merging approach (\S\ref{sec:adapter_merge}), and tests whether Arabic and Persian representations are compatible within a shared parameter space — a prerequisite for merging to succeed.

\subsection{Few-Shot Cross-Lingual Transfer}
\label{sec:fewshot}
%motivation address research question
%description: reproduce experiment with our exlanations
Having established monolingual dapters, we now evaluate few-shot in-context learning (ICL) as the \textbf{first cross-lingual transfer strategy}. The source-language adapter is applied at inference time with target-language demonstrations prepended to the prompt, requiring no additional training. This establishes a lower bound on transfer-learning performance at zero training cost, against which more involved strategies (\S\ref{sec:adapt_ft} and \S\ref{sec:adapter_merge}) can be compared, while also revealing which source adapter provides the strongest transfer signal for each target language.

Three adapter configurations are evaluated: the monolingual Arabic adapter, the monolingual Persian adapter, and the joint AR+FA adapter. Comparing all three reveals whether a single pivot language provides sufficient transfer signal, or whether combining both — either through joint training or merging — is necessary. Both Gemma 2 2B and Llama 3.2 3B are included, since the model inversion hypothesis (that stronger pivot performance does not imply stronger transfer) can only be tested by holding the adapter fixed and varying the base model. Each prompt prepends three target-language examples from FLORES-200, a number small enough to avoid context-window saturation while providing sufficient steering signal \cite{garcia2023unreasonable}.

\subsection{Adaptation Fine-Tuning on Low-Resource Target Languages}
\label{sec:adapt_ft}
%motivation address research question
%description: reproduce experiment with our exlanations
Few-shot ICL (\S\ref{sec:fewshot}) is limited by the number of demonstrations that fit within the context window. We therefore investigate whether directly updating a source adapter on a small target-language dataset yields stronger transfer. Rather than training from scratch, we continue training from our monolingual Arabic and Persian adapters \S\ref{sec:pivot_training}. This warm-start approach is motivated by evidence that minimal parallel data can enable effective multilingual transfer in pre-trained models \cite{wu-etal-2024-far}.

We hypothesise that the source adapter already encodes biomedical domain knowledge from its 25k training pairs, and that a small target-language dataset is sufficient to redirect this knowledge without re-learning the domain. This experiment tests whether minimal adaptation data enables effective cross-lingual transfer, and compares Arabic- versus Persian-initialised adaptation to assess how source–target linguistic relatedness affects transfer.

For each target language, we run two warm-start experiments: one initialised from the Arabic adapter and one from the Persian adapter. This comparison is informative because Arabic and Persian differ in their structural proximity to each target language — if linguistic relatedness drives adaptation efficiency, we expect Persian initialisation to yield stronger results for Dari and Sorani Kurdish, and Arabic initialisation to be more effective for Urdu. The adaptation data is general-domain (FLORES-200), which is a deliberate choice: we are not trying to teach the model new biomedical knowledge, but to redirect already-encoded domain knowledge toward the target language's surface forms. Whether 500 general-domain sentences are sufficient for this redirection — or whether in-domain adaptation data is necessary — is itself a finding this experiment is designed to reveal.

\subsection{LoRA Adapter Merging for Zero-Data Cross-Lingual Transfer}
\label{sec:adapter_merge}
The transfer strategies explored so far require target-language data. We now investigate whether 
high-quality translation can be achieved by merging the weight vectors of the source-language 
adapters (\S\ref{sec:pivot_training}) via post-hoc tensor arithmetic, requiring no additional GPU training. 
Comparing its performance with joint-training and adaptation quantifies the cost of eliminating 
target-language data entirely.

We apply four merging configurations (formalised in \S\ref{subsec:model-merging}): weighted 
averaging with $\alpha \in \{0.3, 0.7\}$, where $\alpha = 0.3$ favours Persian and $\alpha = 0.7$ 
favours Arabic; TIES-Merging with density $k = 0.2$; and DARE with drop rate $p = 0.3$. All four 
are evaluated against the joint-training upper bound (\S\ref{sec:pivot_training}) and supervised 
adaptation (\S\ref{sec:adapt_ft}).

\section{Experimental Setup (further details)}
\label{sec:exp_setup_details}
This section adds a more detailed explanation of the experimental setup described in \ref{sec:exp_setup}.

We use 4-bit nf4 quantisation to reduce GPU memory consumption sufficiently to fit training within a single GPU's VRAM budget. For our Unsloth-supported models (Gemma 2 2B, Llama 3.2 3B, Qwen 2.5 1.5B), Unsloth's optimised training kernels and custom gradient checkpointing are used, reducing VRAM usage by approximately 30\% \cite{han2023unsloth} relative to standard PyTorch and increasing training throughput. These optimisations are mathematically equivalent to standard HuggingFace training and do not affect model weights or outputs. The per-device batch size is 2 with gradient accumulation over 16 steps, yielding the effective batch size of 32 reported in the main text.

During training, the model's EOS token is appended after \texttt{\{target\}}. During inference, \texttt{\{target\}} is left empty and the model generates after the \texttt{\#\#\# \{Language\} translation:} marker; translations are extracted by splitting the decoded output on this marker and retaining the final segment.  Inference uses  \texttt{max\_new\_tokens = 128} and input truncation at 256 tokens.   Inference is run in batches of 64 sentences — a throughput choice only, as greedy decoding is deterministic per sample regardless of batch size. The prompt template used across all experiments is:

\begin{lstlisting}
Translate the below text from English to {Language}. Only output the
final translation in {Language}; do not include any additional text.

### English text:
{source}

### {Language} translation:
{target}
\end{lstlisting}

To monitor training progress, a \texttt{CheckpointMetricsCallback} evaluates BLEU and CHrF++ on the development set at every checkpoint save (approximately every $\lfloor \text{total steps} / 3 \rfloor$ steps). This produces mid-epoch learning curves used for analysis.

\end{document}